\documentclass[acmtog]{acmart}

\usepackage{booktabs} % For formal tables

\usepackage[ruled]{algorithm2e} % For algorithms

\SetAlFnt{\small}
\SetAlCapFnt{\small}
\SetAlCapNameFnt{\small}
\SetAlCapHSkip{0pt}

\acmJournal{TOG}

\copyrightyear{2026}
\acmYear{2026}
\setcopyright{cc}
\setcctype{by}
\acmConference[SIGGRAPH Conference Papers '26]{Special Interest Group on Computer Graphics and Interactive Techniques Conference Conference Papers}{July 19--23, 2026}{Los Angeles, CA, USA}
\acmBooktitle{Special Interest Group on Computer Graphics and Interactive Techniques Conference Conference Papers (SIGGRAPH Conference Papers '26), July 19--23, 2026, Los Angeles, CA, USA}
\acmDOI{10.1145/3799902.3811175}
\acmISBN{979-8-4007-2554-8/2026/07}

\newcommand{\revision}[1]{\textcolor{black}{#1}}
\begin{document}
\title{Pixal3D: Pixel-Aligned 3D Generation from Images}

\author{Dong-Yang Li}
\orcid{0009-0000-6938-3992}
\affiliation{%
  \institution{BNRist, Department of Computer Science and Technology, Tsinghua University}
  \city{Beijing}
  \country{China}
}
\email{ldy23@mails.tsinghua.edu.cn}

\author{Wang Zhao}
\authornote{Project lead.}
\orcid{0000-0001-8925-8574}
\affiliation{%
  \institution{Tencent ARC Lab}
  \city{Beijing}
  \country{China}
}
\email{thuzhaowang@163.com}

\author{Yuxin Chen}
\orcid{0000-0002-7854-1072}
\affiliation{%
  \institution{Tencent ARC Lab}
  \city{Beijing}
  \country{China}
}
\email{chenyux53@163.com}

\author{Wenbo Hu}
\orcid{0000-0001-6082-4966}
\affiliation{%
  \institution{Tencent ARC Lab}
  \city{Shenzhen}
  \country{China}
}
\email{huwenbodut@gmail.com}

\author{Meng-Hao Guo}
\orcid{0000-0002-4128-4594}
\affiliation{%
  \institution{BNRist, Department of Computer Science and Technology, Tsinghua University}
  \city{Beijing}
  \country{China}
}
\email{gmh@tsinghua.edu.cn}

\author{Fang-Lue Zhang}
\orcid{0000-0002-8728-8726}
\affiliation{%
  \institution{Victoria University of Wellington}
  \city{Wellington}
  \country{New Zealand}
}
\email{z.fanglue@gmail.com}

\author{Ying Shan}
\orcid{0000-0001-7673-8325}
\affiliation{%
  \institution{Tencent ARC Lab}
  \city{Shenzhen}
  \country{China}
}
\email{yingsshan@tencent.com}

\author{Shi-Min Hu}
\authornote{Corresponding author.}
\orcid{0000-0001-7507-6542}
\affiliation{%
  \institution{BNRist, Department of Computer Science and Technology, Tsinghua University}
  \city{Beijing}
  \country{China}
}
\email{shimin@tsinghua.edu.cn}

\renewcommand{\shortauthors}{Li et al.}

\begin{abstract}
Recent advances in 3D generative models have rapidly improved image-to-3D synthesis quality, enabling higher-resolution geometry and more realistic appearance. Yet fidelity, which measures pixel-level faithfulness of the generated 3D asset to the input image, still remains a central bottleneck. We argue this stems from an implicit 2D-3D correspondence issue: most 3D-native generators synthesize shape in canonical space and inject image cues via attention, leaving pixel-to-3D associations ambiguous. To tackle this issue, we draw inspiration from 3D reconstruction and propose Pixal3D, a pixel-aligned 3D generation paradigm for high-fidelity 3D asset creation from images. Instead of generating in a canonical pose, Pixal3D directly generates 3D in a pixel-aligned way, consistent with the input view. To enable this, we introduce a pixel back-projection conditioning scheme that explicitly lifts multi-scale image features into a 3D feature volume, establishing direct pixel-to-3D correspondence without ambiguity. We show that Pixal3D is not only scalable and capable of producing high-quality 3D assets, but also substantially improves fidelity, approaching the fidelity level of reconstruction. Furthermore, Pixal3D naturally extends to multi-view generation by aggregating back-projected feature volumes across views. Finally, we show pixel-aligned generation benefits scene synthesis, and present a modular pipeline that produces high-fidelity, object-separated 3D scenes from images. Pixal3D for the first time demonstrates 3D-native pixel-aligned generation at scale, and provides a new inspiring way towards high-fidelity 3D generation of object or scene from single or multi-view images. \revision{Project page: https://ldyang694.github.io/projects/pixal3d/}
\end{abstract}




\begin{teaserfigure}
    \centering
    \includegraphics[width=1.0\linewidth]{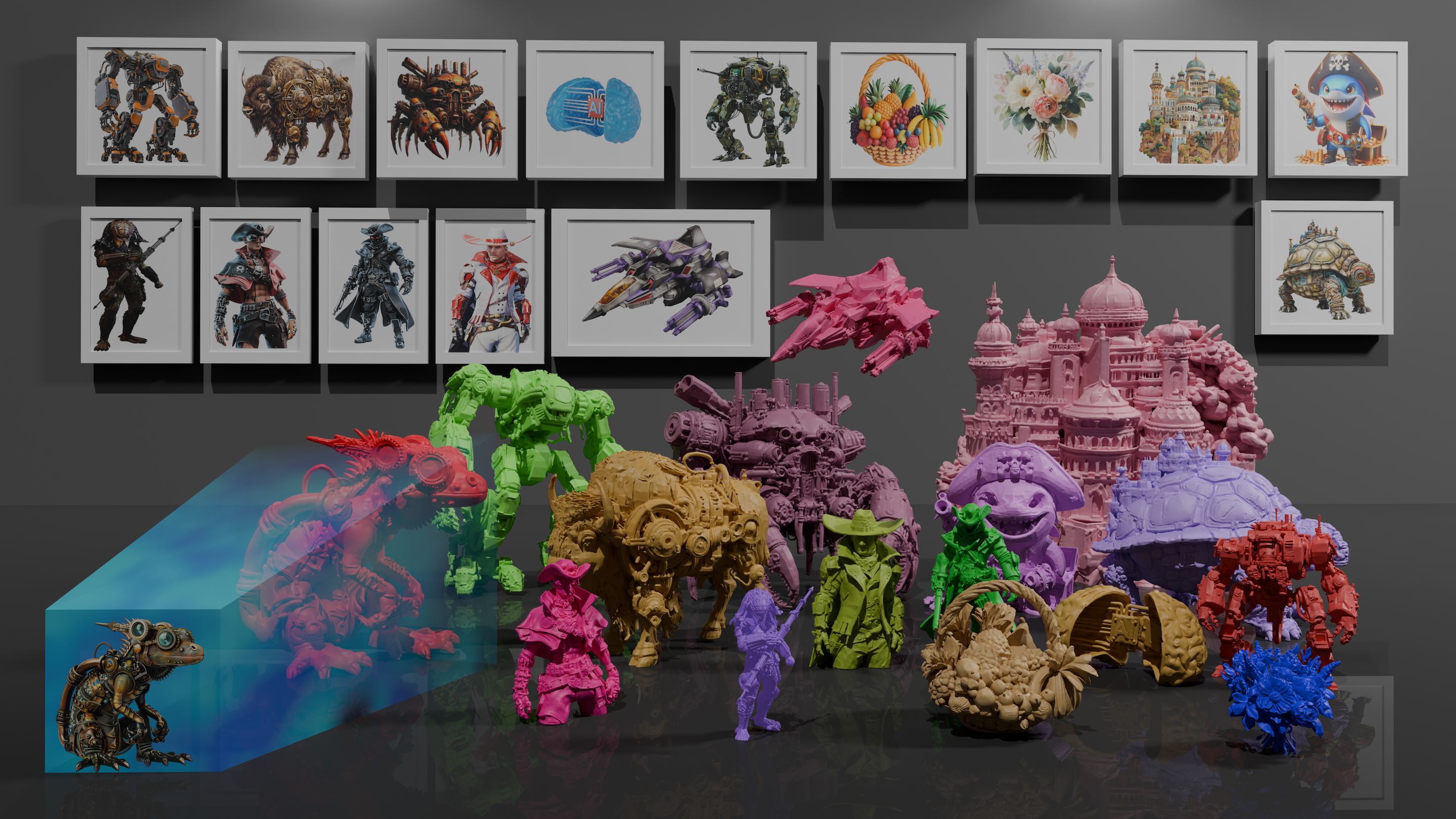}
    \caption{Pixel-aligned meshes generated by Pixal3D. The foreground displays our results with their corresponding input images in the background. Our back-projection conditioning scheme (bottom-left) explicitly lifts 2D image features into a 3D volume to establish robust 2D-3D correspondence for generation.}
    % }
    \label{fig:teaser}
\end{teaserfigure}

\maketitle

\section{Introduction}
Automatic creation of high-quality 3D assets from images is a central goal in computer graphics, with profound implications for gaming, AR/VR, and digital manufacturing. Recent advances in 3D generative modeling have achieved remarkable milestones, producing assets with increasingly detailed geometry~\cite{wu2025direct3d, xiang2025native}, realistic appearance~\cite{yu2024texgen, lai2025natex} and controllable parts~\cite{lin2025partcrafter, yang2025omnipart}, pushing 3D generation towards truly ready-to-use assets.

However, a critical bottleneck still limits the broader adoption of current image-to-3D methods: fidelity. Here, fidelity measures how faithfully the generated 3D asset matches the input image. Most existing methods condition on an image but often produce only approximately similar shapes, with noticeable misalignment and loss of fine details. This falls short of user expectations: given an image, one typically wants the generated 3D model to (1) precisely reconstruct the visible surface, and (2) plausibly complete the unobserved regions to form a coherent and usable 3D asset. Achieving high fidelity, in addition to high quality, is a critical next step towards making image-to-3D generation genuinely useful in practice.

Interestingly, this fidelity issue is far less prominent in 3D reconstruction, a complementary field whose primary goal is to recover visible 3D structure from 2D observations, whether from multiple views or a single view. We attribute this difference to the explicit 2D-3D correspondence establishment. Correspondence is the fundamentals of reconstruction: multi-view geometry~\cite{hartley2003multiple} is built upon pixel correspondences and triangulation, and single-view reconstruction pipelines predict depth~\cite{yang2024depth, lin2025depth}, normals~\cite{fu2024geowizard, ye2024stablenormal}, or point maps~\cite{wang2025moge, szymanowicz2025flash3d} in a pixel-aligned manner, establishing a direct, clear, one-to-one correspondence between 2D image pixels and recovered 3D. In contrast, existing 3D-native generative methods~\cite{xiang2025structured, hunyuan3d2025hunyuan3d, wu2025direct3d} synthesize shapes in a canonical pose, and rely on cross-attention to inject image information into 3D latents. This makes 2D-3D correspondence implicit and nontrivial: cross-attention must effectively "search" for where each image feature should influence the 3D representation, introducing ambiguity and confusion for local details, repetitive parts or among multiple input views, which ultimately manifests as reduced fidelity.

To resolve this fidelity issue, we propose Pixal3D, a new Pixel-Aligned 3D generation paradigm that marries the geometric rigor of reconstruction with the creative power of generative models. Unlike previous canonical space generation, Pixal3D directly generates 3D in a pixel-aligned pose consistent with the input image. To make this possible, we introduce a back-projection conditioning scheme that establishes explicit 2D-3D correspondence for injecting pixel information into 3D, replacing the commonly used cross-attention mechanism. Concretely, we back-project image features into 3D volume: every 3D voxel along that ray is assigned the corresponding pixel feature, yielding a pixel-aligned lifted 3D feature volume. This volume is then added to the 3D noise volume as a conditioning signal. We further incorporate multi-scale image features to preserve and propagate fine-grained details. Through these careful designs, we demonstrate that this pixel-aligned 3D generation paradigm is not only feasible and scalable to produce high-quality 3D models, but also significantly improves 3D fidelity over current 3D generation, achieving near reconstruction-level fidelity.

Moreover, Pixal3D naturally unifies single-view and multi-view settings under the same formulation. We extend Pixal3D to multi-view 3D generation by back-projecting each view into a pixel-aligned feature volume and aggregating them via averaging, leading to a simple and reliable multi-view generation approach. Finally, we show that this pixel-aligned paradigm also benefits 3D scene generation: we propose a modular pipeline that composes object-level generations into high-fidelity, object-separated 3D scenes, in a spirit similar to recent SAM3D~\cite{sam3dteam2025sam3d3dfyimages} scene construction.

Pixal3D is essentially a 3D generative reconstruction paradigm that represents and formalizes the synergy between reconstruction and generation. It inherits the best of both worlds: the visible surfaces are tightly constrained by the input image through explicit correspondence like reconstruction, while the invisible regions are plausibly completed by learned priors of generative model conditioned on what is observed. Pixal3D provides a simple yet effective paradigm for generating faithful 3D objects and scenes from both single-view and multi-view inputs. Figure~\ref{fig:teaser} shows representative examples. Importantly, Pixal3D is orthogonal to specific 3D generative backbones, and can therefore benefit from ongoing advances in geometry representations, part modeling, texturing, materials, etc., making it a scalable foundation for high-fidelity 3D generation.

Our contributions are summarized as follows: (1) We introduce Pixal3D, a pixel-aligned 3D generation paradigm, and demonstrate that pixel-aligned generation is feasible at scale while substantially improving image-to-3D fidelity.
(2) We propose a ray back-projection conditioning mechanism that replaces cross-attention with explicit 2D-3D correspondence, enabling direct pixel-to-3D feature lifting and more faithful preservation of image details.
(3) We extend Pixal3D from single-view to multi-view generation via simple and effective multi-view feature-volume aggregation.
(4) We propose a modular 3D scene generation pipeline based on Pixal3D that produces high-fidelity, object-separated 3D scenes.

\section{Related Works}
\subsection{3D Generation}
3D generation has advanced rapidly~\cite{wang2025diffusion3dsurvey}, from distilling 2D diffusion priors into 3D~\cite{poole2022dreamfusion, wang2023prolificdreamer} to 3D-native pipelines that learn 3D distributions from large-scale datasets~\cite{deitke2023objaversexl}. A key driver is designing 3D representations that balance fidelity, efficiency, and scalability, spanning point clouds~\cite{nichol2022point}, voxels~\cite{xiong2025octfusion}, meshes~\cite{liu2023meshdiffusion}, 3D Gaussians~\cite{yushi2025gaussiananything}, and triplanes~\cite{wu2024direct3d}, etc. 3DShape2VecSet \cite{zhang20233dshape2vecset} introduced latent vector sets as implicit representation, later adopted and extended by~\cite{zhang2024clay, li2025craftsman3d, zhao2025hunyuan3d, li2025triposg,li2025relate3d} to demonstrate its scalability. To relief fidelity issue, Hi3DGen~\cite{ye2025hi3dgen} introduced normal as both input and regularization. TRELLIS~\cite{xiang2025structured} proposed a sparse voxel unified representation for jointly embedding geometry and appearance, and Direct3D-S2~\cite{wu2025direct3d} improved sparse voxel efficiency and regularity via spatial sparse attention. Flexible and deformable surface parameterizations are explored in Sparc3D~\cite{li2025sparc3d} and TripoSF~\cite{he2025sparseflex}, enabling the generation of intricate structures and open surfaces. Inspired by Dual Contouring~\cite{ju2002dual}, TRELLIS 2~\cite{xiang2025native} and FaithC~\cite{luo2025faithful} incorporated dual-grid information to enhance surface representation quality. LATTICE~\cite{lai2025lattice} combined compact vector sets with structural sparse voxels, proposing VoxSet for scalable generation.

Despite this progress, current state-of-the-art image-to-3D generation still faces a well-known fidelity issue: outputs are often not pixel-faithful to the input image as in reconstruction. Notably, all above methods create 3D shapes in canonical poses and condition images via cross-attention, leaving 2D-3D correspondence implicit and ambiguous, which we argue is a key cause of reduced fidelity. In contrast, Pixal3D explores a new generation paradigm to directly generate pixel-aligned 3D objects, demonstrating superior fidelity while remaining compatible with the above representation and architectural advances.

\subsection{3D Reconstruction}
3D reconstruction from images is a long-standing visual problem. Classical structure-from-motion (SfM) and multi-view stereo (MVS) \cite{schoenberger2016sfm, schoenberger2016mvs} recover 3D structure by establishing correspondences, triangulation, and 2D-3D optimization such as bundle adjustment. With deep learning, approaches~\cite{huang2018deepmvs, yao2018mvsnet, im2019dpsnet} explored plane-sweeping of deep features to improve MVS robustness. Beyond 2.5D, Atlas~\cite{murez2020atlas} back-projects image features into a voxel grid for direct 3D prediction with 3D CNNs, and NeuralRecon~\cite{sun2021neuralrecon} extends this to streaming reconstruction with similar back-projection. Our Pixal3D is inspired by these pioneers and integrates pixel-aligned back-projection into a generative backbone. Recently, feed-forward multi-view reconstruction methods like DUSt3R~\cite{wang2024dust3r}, VGGT~\cite{wang2025vggt} and their followers~\cite{tang2025mv, yang2025fast3r} have shown strong scalability by predicting pixel-aligned point maps in a shared coordinate. Similarly, single-image reconstruction has advanced including depth~\cite{yang2024depth, yin2023metric3d, ke2024repurposing, meng2025indoor3dsurvey, lin2025depth}, normal~\cite{hu2024metric3d, ye2024stablenormal, fu2024geowizard}, point map~\cite{wang2025moge, wang2025moge2} or 3D Gaussian~\cite{szymanowicz2025flash3d, szymanowicz2025bolt3d,zheng2024gpsgaussian} prediction in a pixel-aligned manner.

While reconstruction recovers visible surfaces with high fidelity, its outputs are incomplete and thus not directly usable as 3D assets. Nevertheless, the explicit and unambiguous 2D-3D correspondence in reconstruction provides a key insight for generation. Pixal3D brings this principle to 3D generation via pixel-aligned modeling, enabling complete asset creation with reconstruction-level fidelity.

\subsection{3D Generative Reconstruction}
As 3D reconstruction and 3D generation mature, researchers increasingly realize their complementarity. This gives rise to 3D generative reconstruction, which couples reconstruction constraints with generative modeling to obtain outputs that are both consistent with inputs and complete/plausible beyond them. Early works used image generative model to complete insufficient 2D views~\cite{shi2023mvdream, liu2023zero} to enhance reconstruction~\cite{hong2023lrm, li2023instant3d}. RaySt3R~\cite{duisterhof2025rayst3r} performs ray-based novel-view prediction and fuses multi-view estimates into a complete shape, while Gen3R~\cite{huang2026gen3r} couples a feed-forward reconstruction backbone with diffusion to align geometry and appearance. LaRI~\cite{li2025lari} introduces view-aligned layered ray-intersection representations to better reason over occlusions. Closest to our motivation, recent works ReconViaGen~\cite{chang2025reconviagen} and CUPID~\cite{huang2025cupid} target high-fidelity generative reconstruction. ReconViaGen injects VGGT features into a canonical-space generator, and CUPID jointly models a canonical 3D object and camera pose. In contrast, Pixal3D pushes this integration further and thoroughly, by establishing and enforcing explicit 2D-3D correspondence rather than predicting it: we directly generating 3d object in a pixel-aligned view-centric manner via back-projection. This design avoids the brittleness of camera estimation and reduces fidelity loss introduced by canonical-pose generation and predicted-pose dependent pixel feature fetching, leading to a scalable foundation for 3D generative reconstruction.
\begin{figure*}[t]
\centering
% \vskip -0.3in
\includegraphics[width=0.95\textwidth]{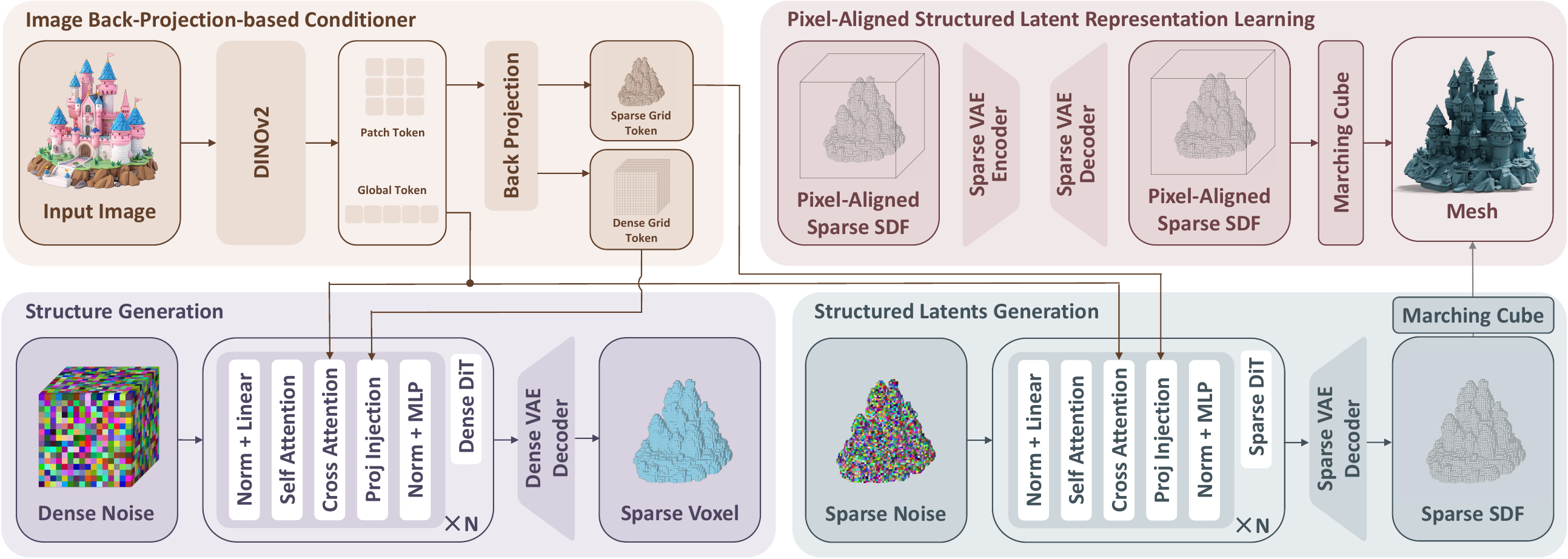}
\vskip -0.1in
\caption{
Overview of the Pixal3D framework. The framework consists of three key components: (1) Pixel-Aligned Structured Latent Representation Learning (top-right), which uses a VAE to compress pixel-aligned sparse SDF into efficient sparse latents; (2) an Image Back-Projection-based Conditioner (top-left) that explicitly lifts 2D image features into 3D feature volumes; and (3) a two-stage generative process (Structure Generation and Structured Latents Generation) conditioned on these volumes to predict coarse structure and detailed latents, respectively. Finally, the generated latents are decoded into a high-fidelity mesh.}
\label{fig:pipeline}
\vskip -0.1in
\end{figure*}

\section{Method}
Pixal3D introduces a pixel-aligned 3D generation paradigm and proposes a back-projection-based image condition scheme into a 3D latent diffusion model. This paradigm is further extended to support multi-view generation and modular scene-level synthesis. An overview of the framework is shown in Figure~\ref{fig:pipeline}. Next, we first summarize the preliminaries of our base 3D latent diffusion model in Sec.~\ref{sec:pre}, then detail the pixel-aligned 3D generation in Sec.~\ref{sec:gen}, present the modular scene generation pipeline in Sec.~\ref{sec:scene}, and discuss implementation details in Sec.~\ref{sec:imple}.

\subsection{Preliminary}
\label{sec:pre}
In principle, Pixal3D is compatible with any explicitly structured 3D generation backbone. In this work, we adopt the open-source state-of-the-art model Direct3D-S2~\cite{wu2025direct3d} as our base. Direct3D-S2 is a 3D latent diffusion framework utilizing sparse voxel latents as its 3D representation. Similar to TRELLIS~\cite{xiang2025structured}, it consists of a dense stage and a sparse stage, each equipped with its own VAE and DiT model. The dense stage encodes and samples a coarse occupancy grid, which is used to determine the voxel indices for the subsequent sparse stage. In the sparse stage, a sparse DiT denoises noisy sparse voxel latents, which are then decoded by a VAE decoder into a sparse SDF. Applying Marching Cubes subsequently yields the final mesh. In both the dense and sparse DiT models, image conditioning is injected via cross-attention. Pixal3D retains the core architecture of Direct3D-S2 and extends it by introducing a pixel-aligned generation paradigm.

\subsection{Pixel-aligned 3D Generation}
\label{sec:gen}

\subsubsection{Canonical vs. Pixel-Aligned Generation:}
Existing 3D-native generation methods typically operate in an object-centric canonical pose. This representation defines a default, view-independent orientation for an object, anchoring its semantic components (e.g., a car's front, a chair's seat) to predefined axes. While this paradigm facilitates learning robust category-level priors, it fundamentally underconstrains the 2D-3D correspondence for image-conditioned generation. In practice, this correspondence is established through cross-attention between 2D and 3D tokens as a \textit{learned behavior}. This process is inherently ambiguous: multiple 3D locations in canonical space can explain similar 2D evidence under unknown pose. Consequently, the model often cheats by using global semantic cues rather than establishing a mathematically faithful pixel-to-3D mapping.

In contrast, Pixal3D introduces \textit{pixel-aligned generation}, where objects are defined in the input camera’s coordinate frame. Intuitively, the object is represented "as seen from the camera". The generator builds view-dependent 3D behind pixels: the 3D volume is aligned with the image frustum, so each pixel corresponds to a unique camera ray and therefore a structured locus in 3D. This alignment turns correspondence from a learned, stochastic behavior into a solid geometric prior. Next, we introduce our back-projection conditioned 3D latent diffusion to realize this pixel-aligned 3D generation.

\subsubsection{Back-projection Conditioned 3D Latent Diffusion.}
Pixal3D is built upon 3D latent diffusion models, as introduced in Sec.~\ref{sec:pre}. Unlike existing methods, where structured latents encoded from canonical objects serve as the diffusion target, our VAE model encodes pixel-aligned objects into 3D latents, as shown in Figure~\ref{fig:pipeline}. Different input views thus correspond to different camera-space objects $\mathbf{X}$ and thus different latents $\mathbf{z}_0$. The diffusion model therefore learns view-dependent, pixel-aligned generation.

\begin{figure}[t!]
\centering

\includegraphics[width=0.45\textwidth]{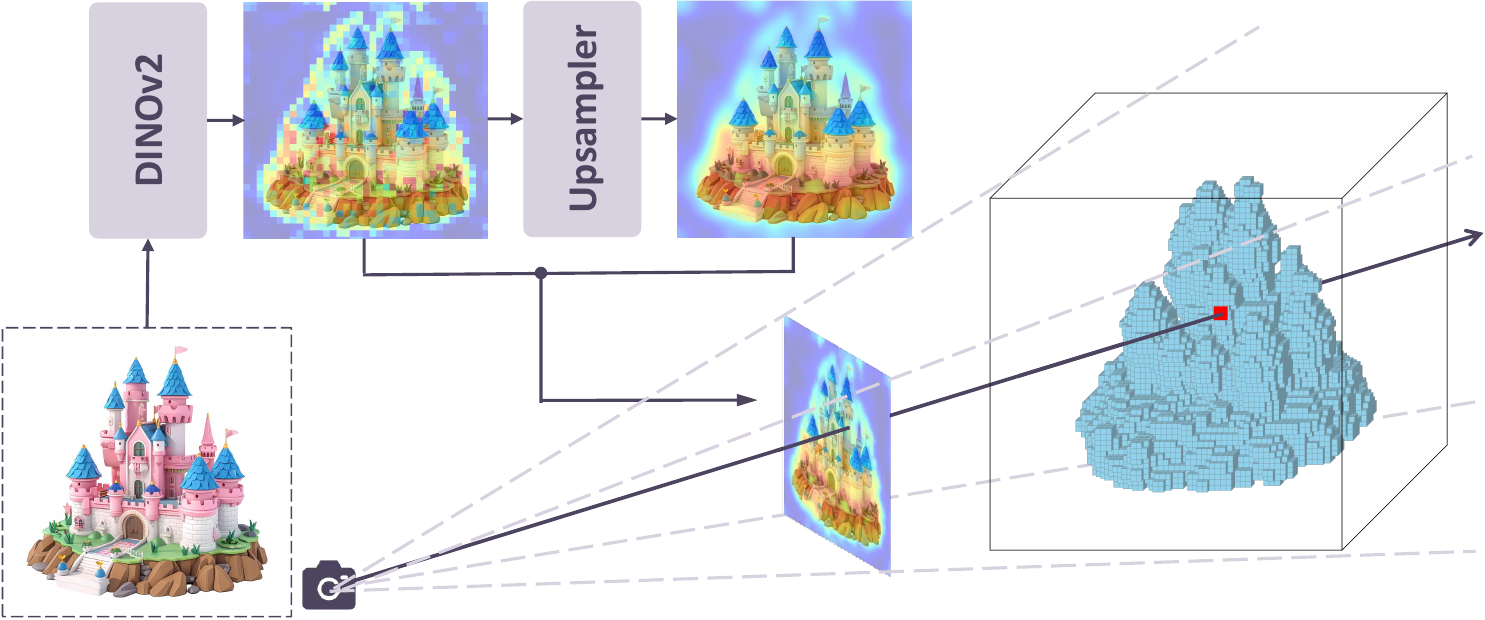}
\vskip -0.1in
\caption{Illustration of the Back-projection Conditioning Scheme.
}
\label{fig:projection}
\vskip -0.2in
\end{figure}

\paragraph{Back-projection condition scheme.} To enable pixel-aligned generation, we introduce a back-projection scheme, instead of cross-attention, for injecting 2D image information into 3D, as shown in Figure~\ref{fig:projection}. Specifically, given an input image $I$, we first extract a 2D feature map $I^{'}$ using DINOv2~\cite{oquab2023dinov2}. Each pixel in this feature map can be back-projected into a ray within the 3D camera coordinate system. Any 3D point along such a ray represents a potential surface point of the target object. Collectively, these rays form a camera frustum, within which the target 3D shape is assumed to reside and be defined by the image-conditioned rays.

% \begin{figure*}[tb]
%    \vskip -0.1in
%     \newcommand{\signleviewcoloredrow}[1]{
%        \includegraphics[width=0.16\linewidth]{figures/images/singleview_figure_image/#1_input_single_view.png} &
%     \includegraphics[width=0.16\linewidth]{figures/images/singleview_figure_image/#1_trellis_colored_single_view.png} &
%     \includegraphics[width=0.16\linewidth]{figures/images/singleview_figure_image/#1_triposg_colored_single_view.png} &
%         \includegraphics[width=0.16\linewidth]{figures/images/singleview_figure_image/#1_hunyuan3d_colored_single_view.png} &
%     \includegraphics[width=0.16\linewidth]{figures/images/singleview_figure_image/#1_direct3ds2_colored_single_view.png} &
%         \includegraphics[width=0.16\linewidth]{figures/images/singleview_figure_image/#1_pa3d_single_view.png} 
%         \\
%     }
%     \newcommand{\singleviewmainrow}[1]{\signleviewcoloredrow{#1}}
    
%     \centering
%     \small
%     \setlength{\tabcolsep}{0pt}
%     \begin{tabular}{cccccc}

%         \centering

%         \singleviewmainrow{81}

%         \singleviewmainrow{150}
   
%        Input & TRELLIS & TripoSG & Hunyuan3D-2.1 & Direct3D-S2 & Pixal3D  \\
%     \end{tabular}
%     \vskip -0.1in
%   \caption{Qualitative comparisons of single view 3D generation.}
%    \vskip -0.1in
%     \label{fig:singleiview_main}
% \end{figure*}
\begin{figure*}[tb]
   \vskip -0.1in
    
    \centering
    \small
    \setlength{\tabcolsep}{0pt}
    \begin{tabular}{cccccc}

        \centering
  
        \includegraphics[width=0.16\linewidth]{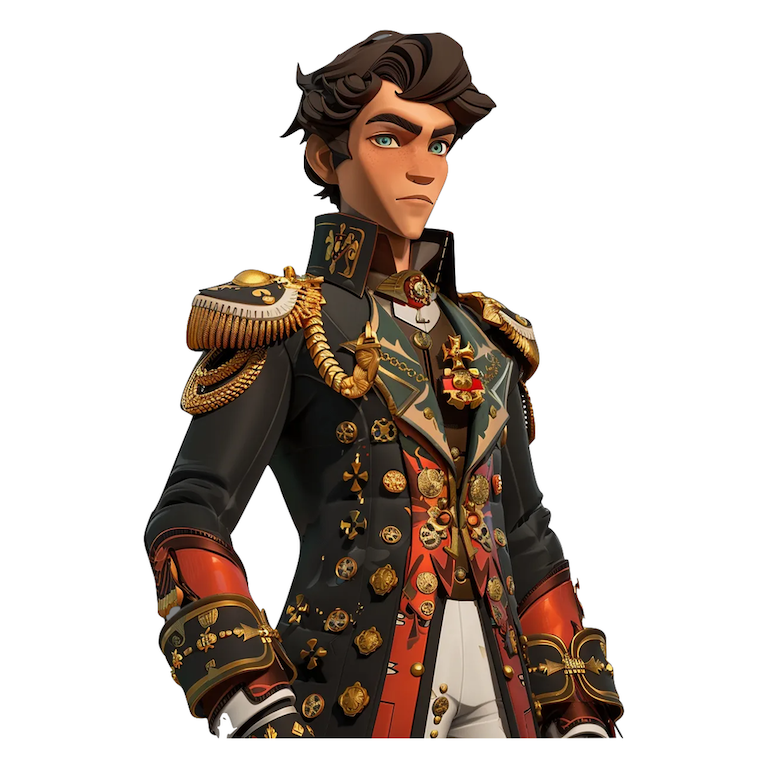} &
        \includegraphics[width=0.16\linewidth]{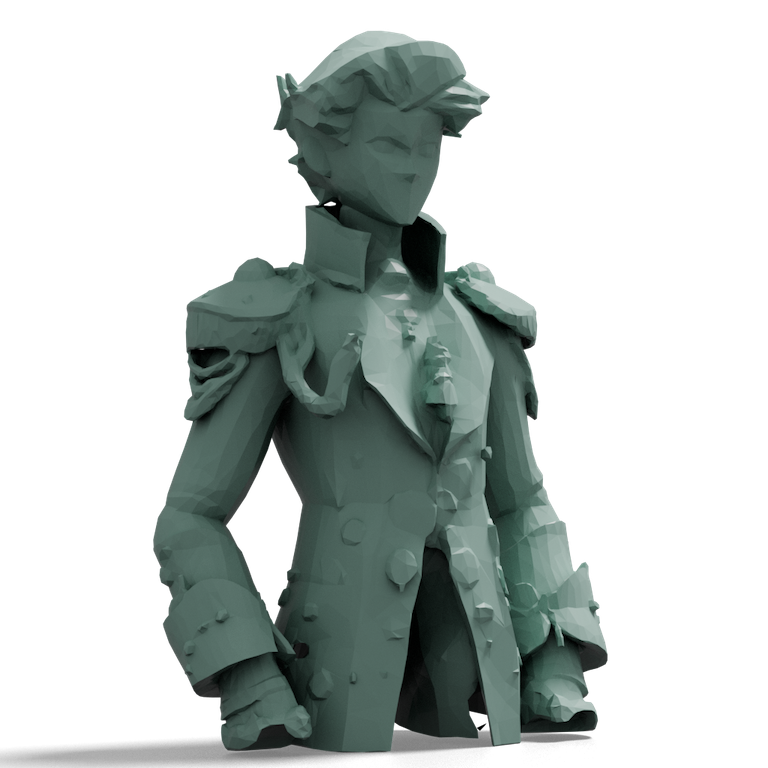} &
        \includegraphics[width=0.16\linewidth]{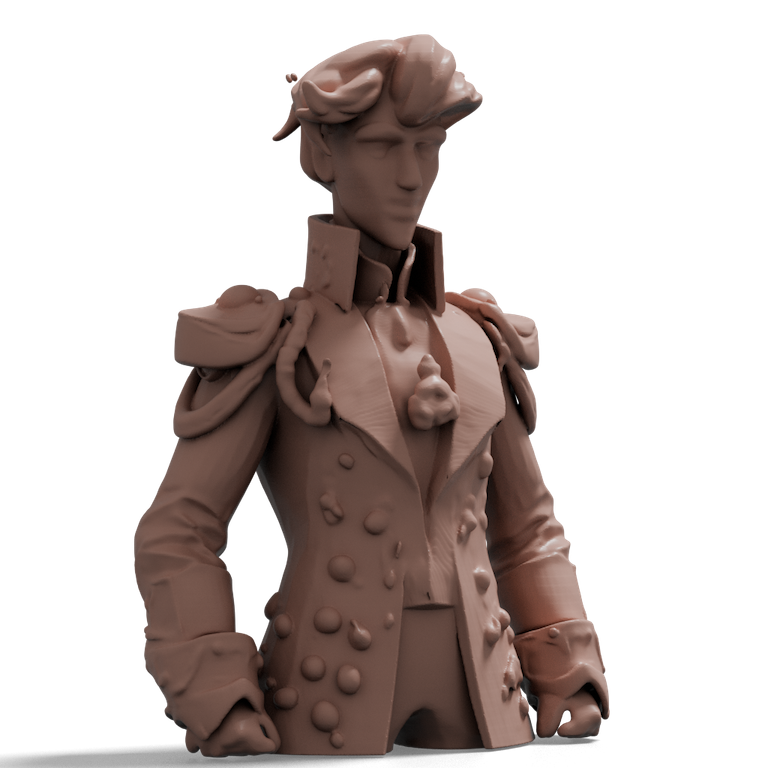} &
        \includegraphics[width=0.16\linewidth]{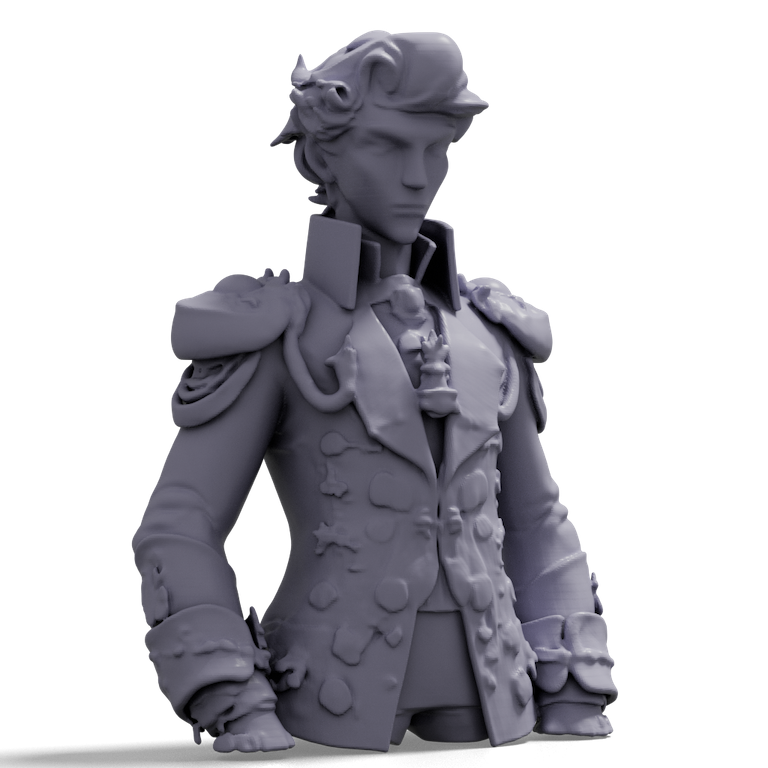} &
        \includegraphics[width=0.16\linewidth]{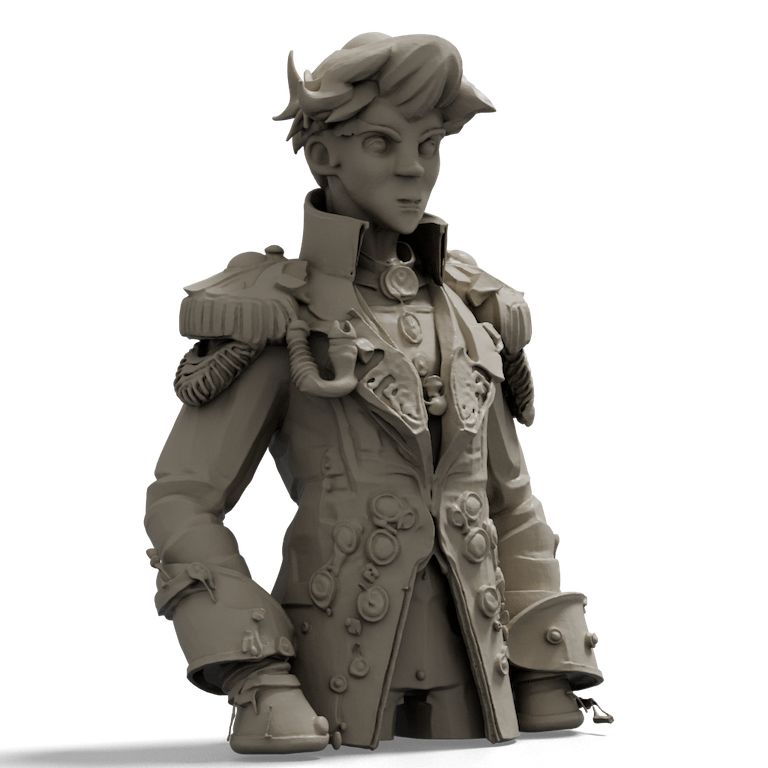} &
        \includegraphics[width=0.16\linewidth]{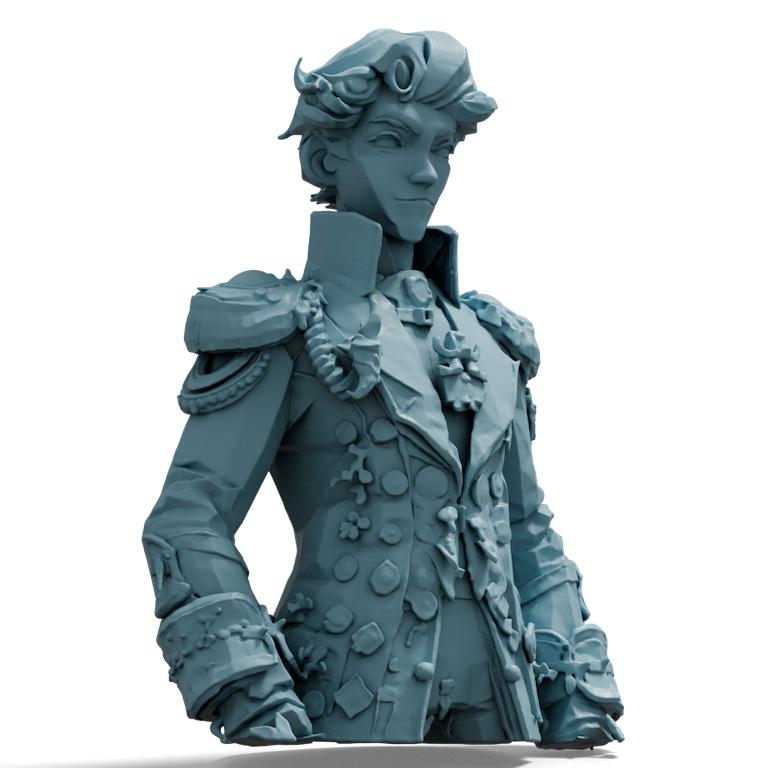} \\

        \includegraphics[width=0.16\linewidth]{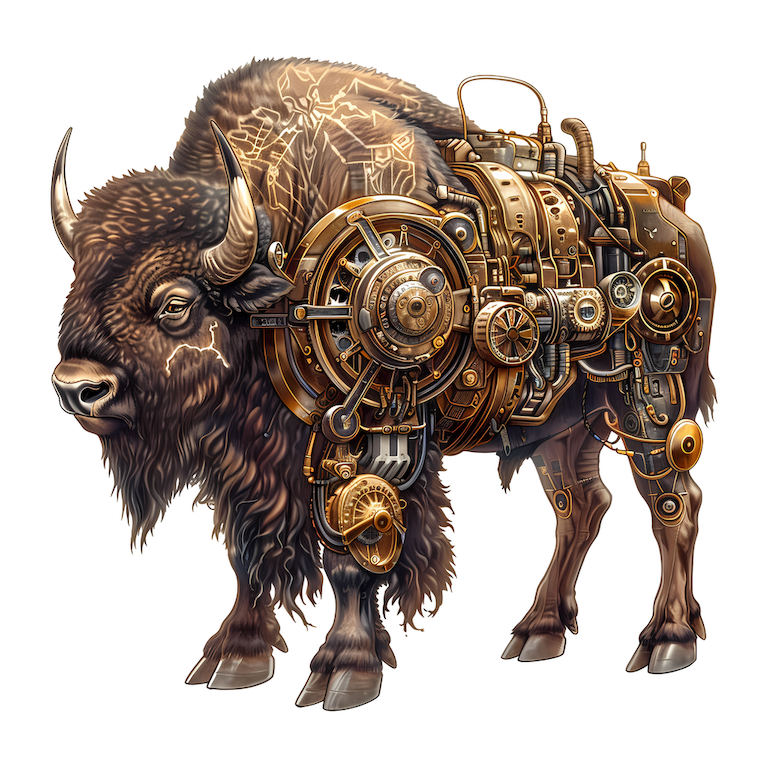} &
        \includegraphics[width=0.16\linewidth]{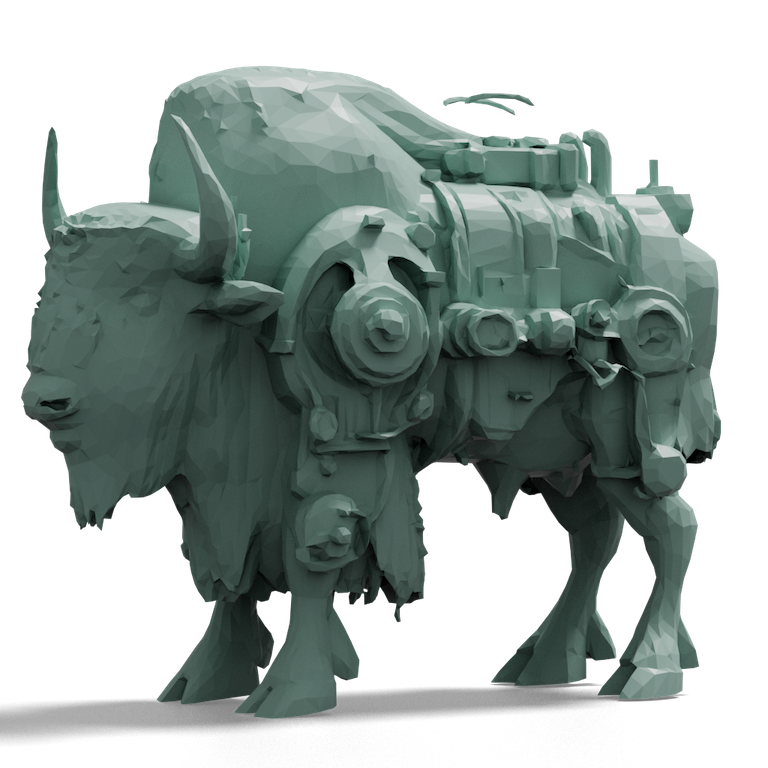} &
        \includegraphics[width=0.16\linewidth]{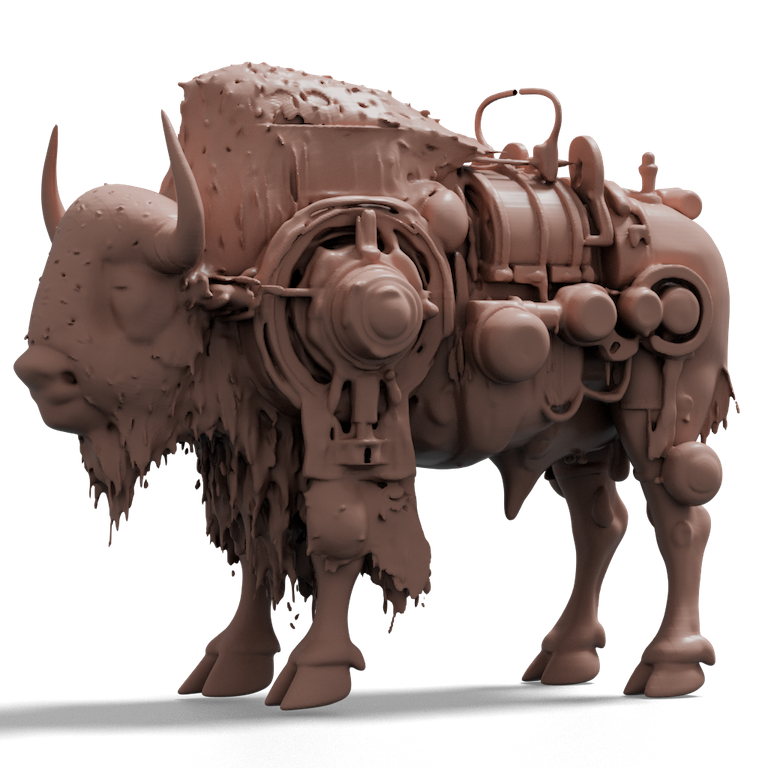} &
        \includegraphics[width=0.16\linewidth]{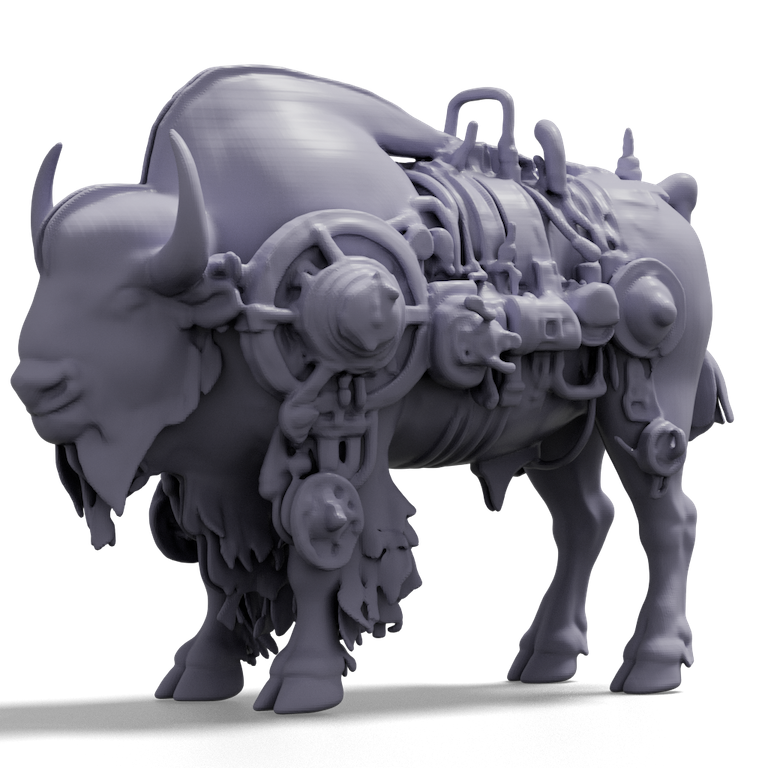} &
        \includegraphics[width=0.16\linewidth]{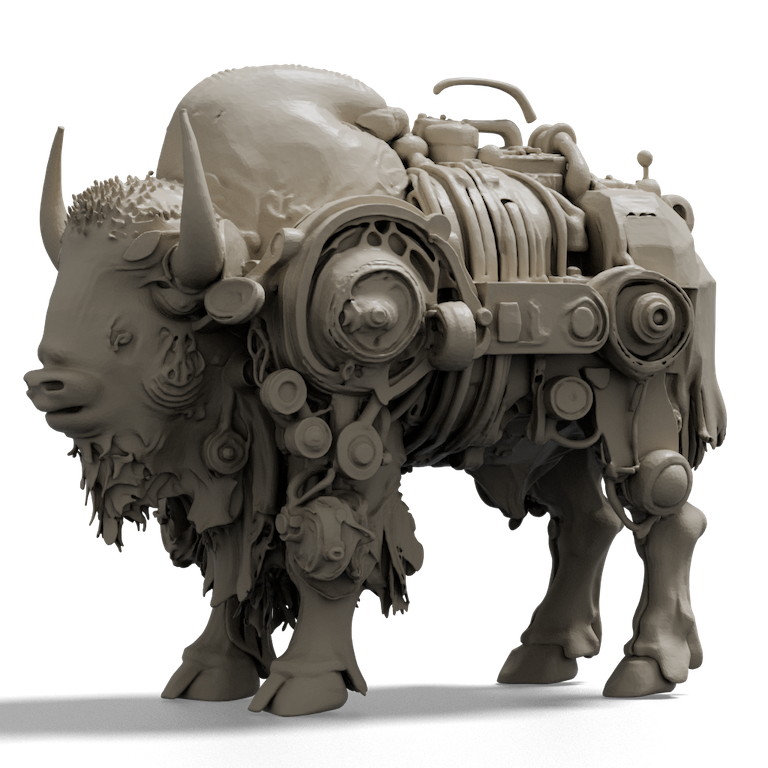} &
        \includegraphics[width=0.16\linewidth]{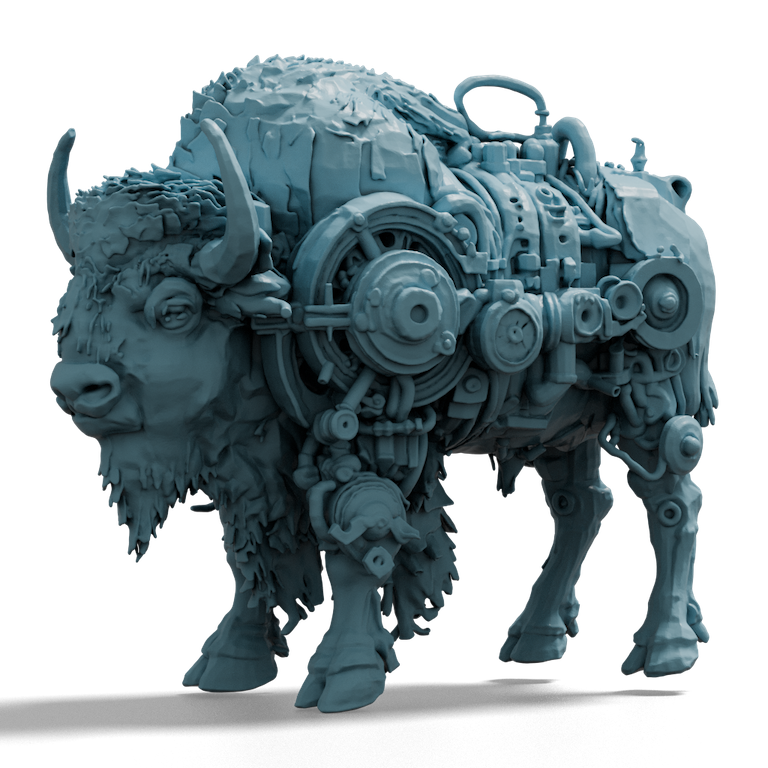} \\
   
        Input & TRELLIS & TripoSG & Hunyuan3D-2.1 & Direct3D-S2 & Pixal3D  \\
    \end{tabular}
    \vskip -0.1in

    \caption{Qualitative comparisons of single view 3D generation.}
    \vskip -0.1in
    \label{fig:singleiview_main}
\end{figure*}

The object can theoretically exist at any scale along this frustum, similar to the scale ambiguity in single-view depth estimation. However, 3D generative models often require a predefined bounding box, typically a unit cube, to specify the normalized spatial extent of the object. This cube is then voxelized (e.g., at a $64^3$ resolution) to serve as input for the generative model. Therefore, we need to determine what size the cube is and where this cube should be placed within the camera frustum. We aim to ensure that the cube is not so large that the projected rays occupy only a small fraction of the voxels (degrading resolution and efficiency), yet not so small that it fails to capture the full extent of the frustum (leading to information loss). This placement is governed by a distance parameter $d$, which represents the distance from the camera plane to the center of the cube, and a cube scale parameter $s$ that controls the size of the cube. With these parameters, an explicit 2D-3D correspondence can be established between image pixel $(u, v)$ and voxel $(i, j, k)$ inside the cube through the projection formula.

In this manner, each voxel gathers image features from its corresponding ray, forming a 3D feature volume. This feature volume provides pixel-aligned image information, which the 3D generative model uses for sampling and generation. In practice, the above process is implemented in the reverse direction following previous methods~\cite{murez2020atlas, sun2021neuralrecon}: we project voxels onto the image plane and sample features from the image, which makes it simpler and more effective to handle interpolation. During training, we use ground-truth projection parameters, including camera intrinsics, distance $d$ and cube scale $s$. For inference, we do not require these parameters; instead, we select a relatively small field of view, a unit cube scale, and then compute the camera distance such that the rays cast from the four image corners pass exactly through the four vertices of the back face of the unit cube. This ensures that the frustum information inside the cube is complete, while not sacrificing too much voxel utilization. In practice, this strategy is stable and robust, and we adopt it for all subsequent experiments.

The resulting feature volume is spatially aligned with the noise volume in the diffusion model. Therefore, we directly add the feature volume to the noise volume as the image condition. Meanwhile, we also inject the global feature token extracted by DINOv2 (image-level rather than patch-level, originally used for classification) via cross-attention, providing additional global semantic guidance.

\paragraph{Multi-scale 2D feature maps.} While DINOv2 features contain rich image information, they are primarily composed of high-level semantic features with relatively coarse granularity. Consequently, low-level, fine-grained structural details are often lost, which to some extent constrains the fidelity of image-to-3D generation. To address this, we propose leveraging multi-scale image features to simultaneously preserve both low-level and high-level information.

Specifically, we use an off-the-shelf feature upsampling model \cite{chambon2025naf} to upscale the DINOv2 patch-token features $I'$ to full resolution, producing a detail-rich, image-consistent map $I^h$. Our back-projection conditioning is unchanged: we project each voxel to the image plane, bilinearly sample features at each scale, and average them. These multi-scale high-resolution features improve the recovery and consistency of fine details. Also, it highlights the advantage of our pixel-aligned paradigm: unlike cross-attention-based methods where dense attention to high-resolution maps is prohibitively expensive, our explicit 2D-3D correspondence makes this upgrade essentially cost-free while yielding measurable gains.

\subsubsection{Multi-view Extension.}
Since our single-view model is formulated explicitly through projection geometry, extending it to the multi-view setting is naturally straightforward. Unlike the single-view case, we assume that the camera parameters for all input views are known, consistent with the established standard setups in traditional multi-view stereo, NeRF, and 3D Gaussian Splatting, etc. Given a set of multi-view images, we back-project the multi-scale features from each view into 3D space and aggregate them within each voxel by simple averaging. The resulting fused feature volume is then used as the conditioning signal for the generative model. This simple yet effective multi-view conditioning scheme accommodates an arbitrary number of input views. As the number of viewpoints increases, a greater extent of the 3D surface becomes visible, leading to a more deterministic 3D shape reconstruction. 

% \begin{figure*}[!p]
%     \newcommand{\signleviewcoloredrow}[1]{
%        \includegraphics[width=0.16\linewidth]{figures/images/singleview_figure_image/#1_input_single_view.png} &
%     \includegraphics[width=0.16\linewidth]{figures/images/singleview_figure_image/#1_trellis_colored_single_view.png} &
%     \includegraphics[width=0.16\linewidth]{figures/images/singleview_figure_image/#1_triposg_colored_single_view.png} &
%         \includegraphics[width=0.16\linewidth]{figures/images/singleview_figure_image/#1_hunyuan3d_colored_single_view.png} &
%     \includegraphics[width=0.16\linewidth]{figures/images/singleview_figure_image/#1_direct3ds2_colored_single_view.png} &
%         \includegraphics[width=0.16\linewidth]{figures/images/singleview_figure_image/#1_pa3d_single_view.png} 
%         \\
%     }
%     \newcommand{\singleviewmainrow}[1]{\signleviewcoloredrow{#1}}
    
%     \centering
%     \small
%     \setlength{\tabcolsep}{0pt}
%     \begin{tabular}{cccccc}

%         \centering
  
%         \singleviewmainrow{52}
%         \singleviewmainrow{79}
%       \singleviewmainrow{126}
%          \singleviewmainrow{141}
%         \singleviewmainrow{140}
%         \singleviewmainrow{157}
%         \singleviewmainrow{102}
%        Input & TRELLIS & TripoSG & Hunyuan3D-2.1 & Direct3D-S2 & Pixal3D  \\
%     \end{tabular}
%   \caption{Qualitative comparison of single-view 3D generation on in-the-wild images. }
%     \label{fig_only:singleiview}
% \end{figure*}
\begin{figure*}[!p]
    
    \centering
    \small
    \setlength{\tabcolsep}{0pt}
    \begin{tabular}{cccccc}

        \centering
  
        \includegraphics[width=0.16\linewidth]{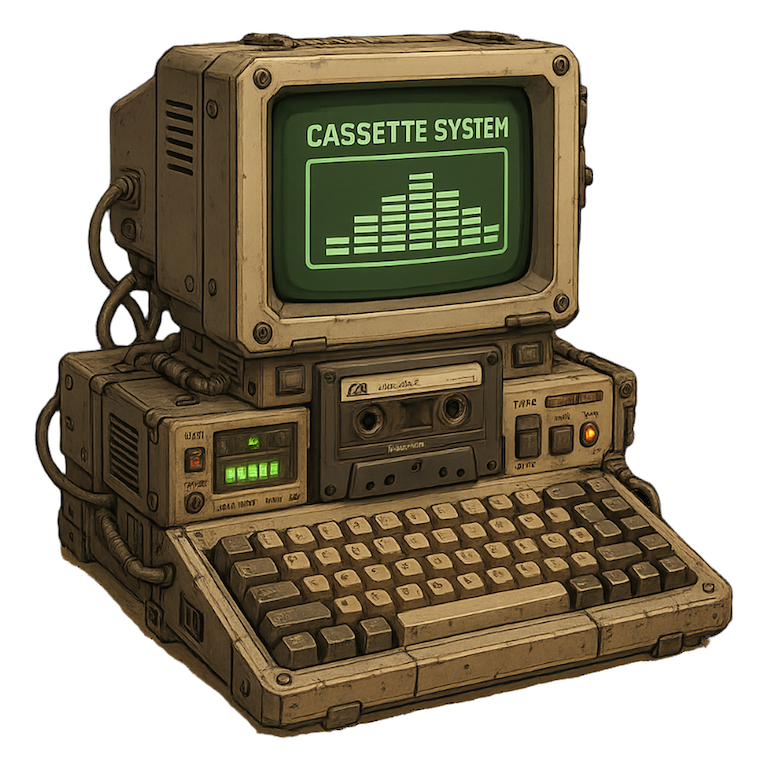} &
        \includegraphics[width=0.16\linewidth]{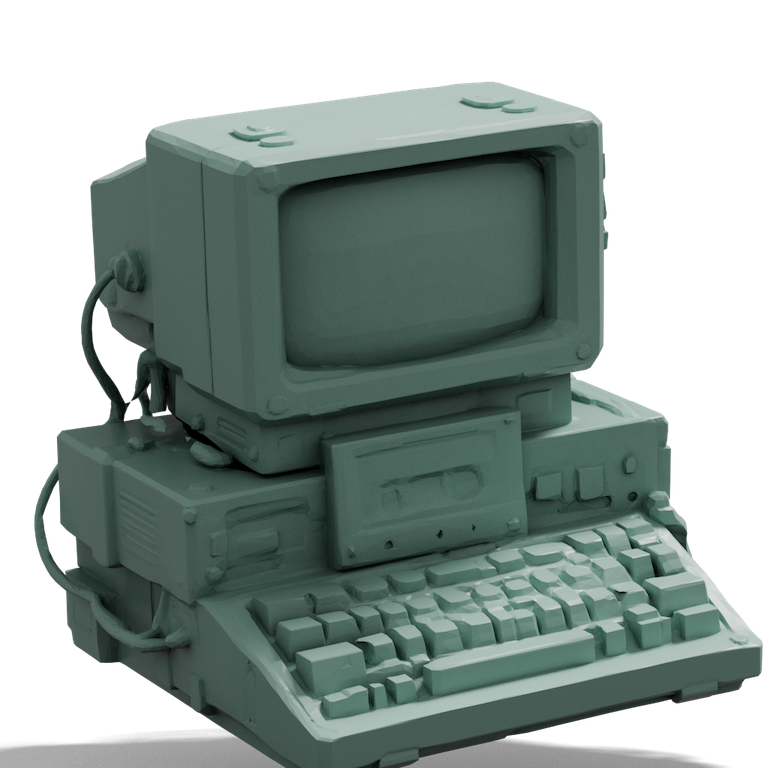} &
        \includegraphics[width=0.16\linewidth]{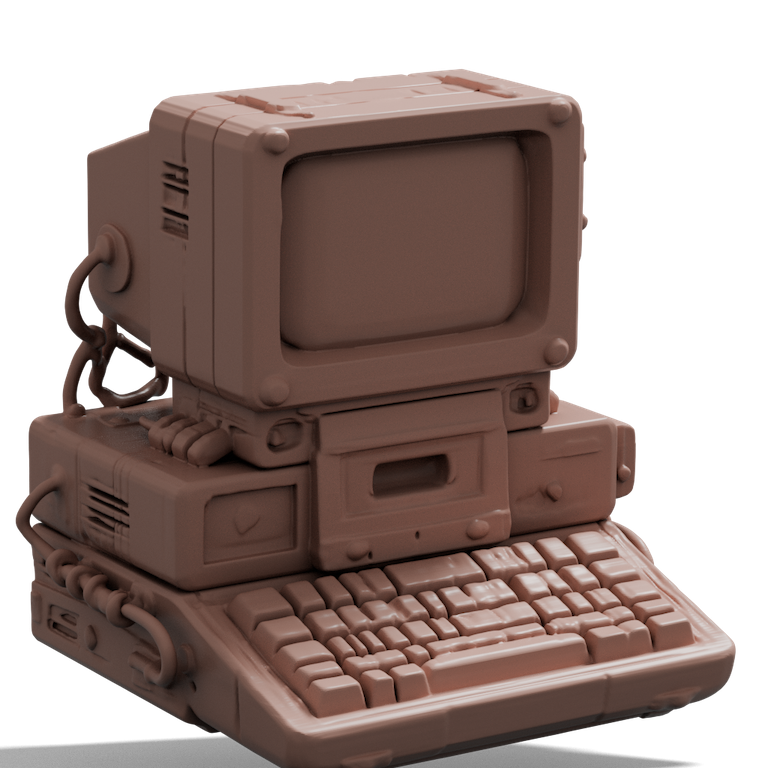} &
        \includegraphics[width=0.16\linewidth]{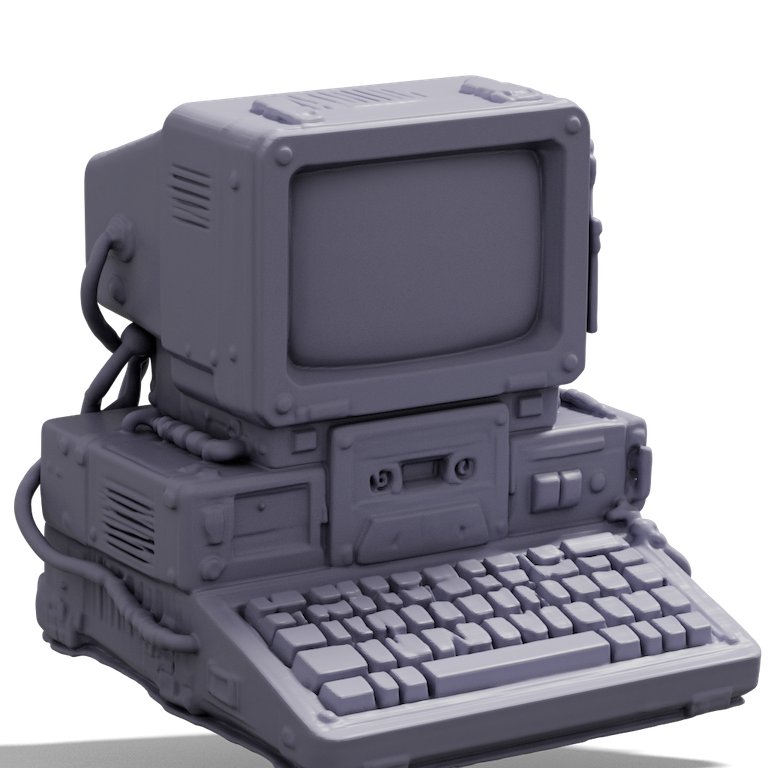} &
        \includegraphics[width=0.16\linewidth]{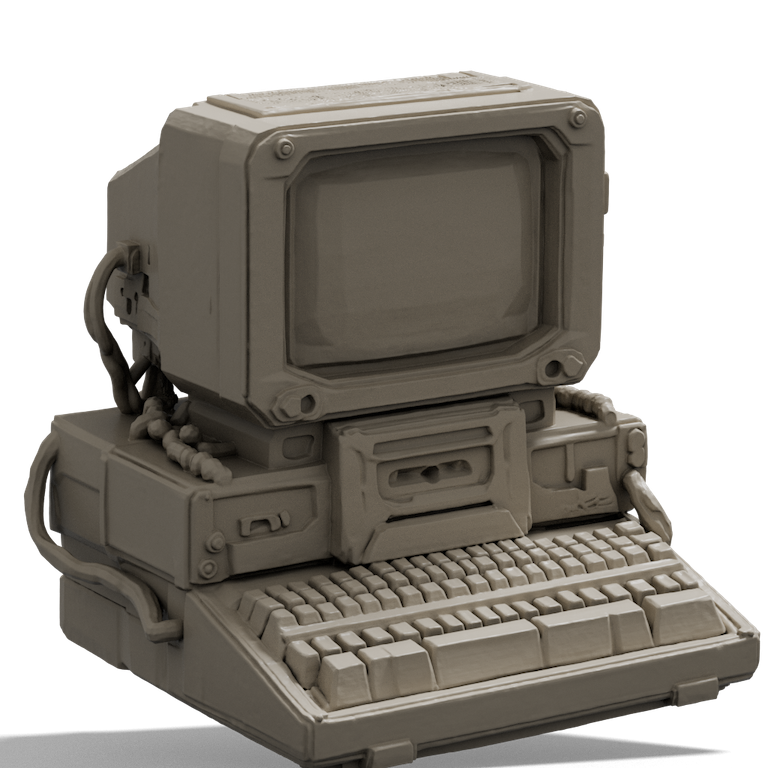} &
        \includegraphics[width=0.16\linewidth]{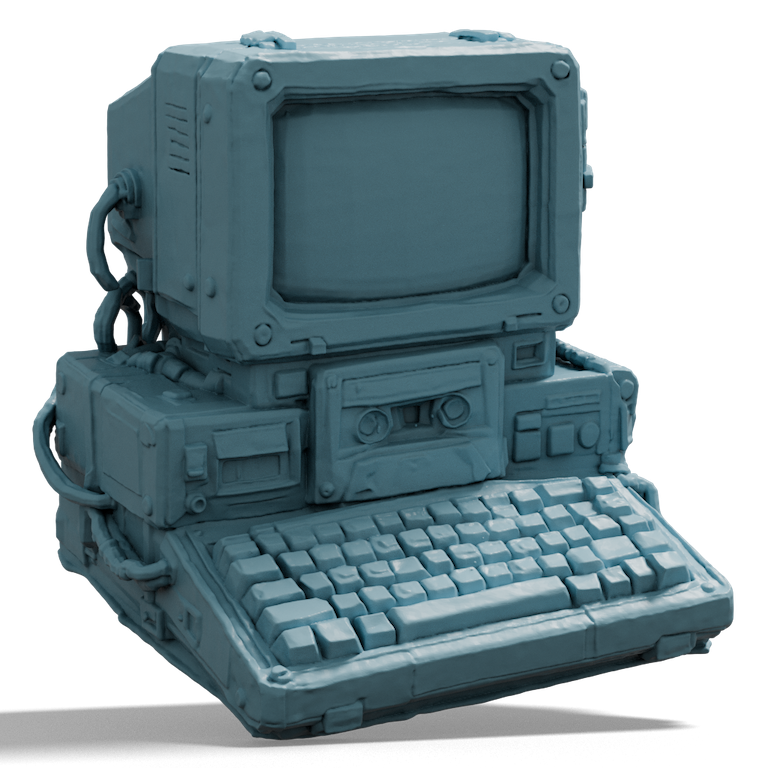} \\

        \includegraphics[width=0.16\linewidth]{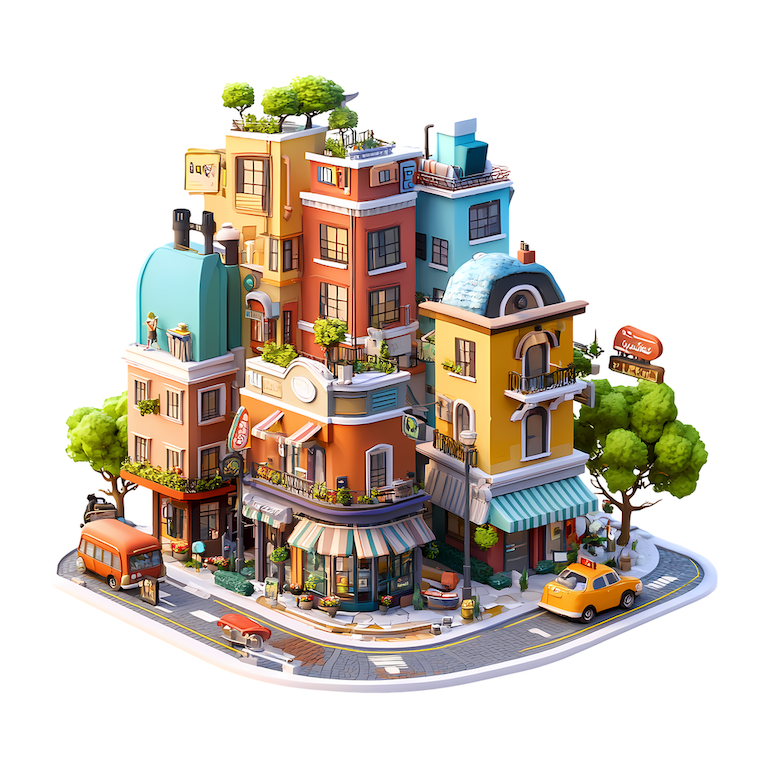} &
        \includegraphics[width=0.16\linewidth]{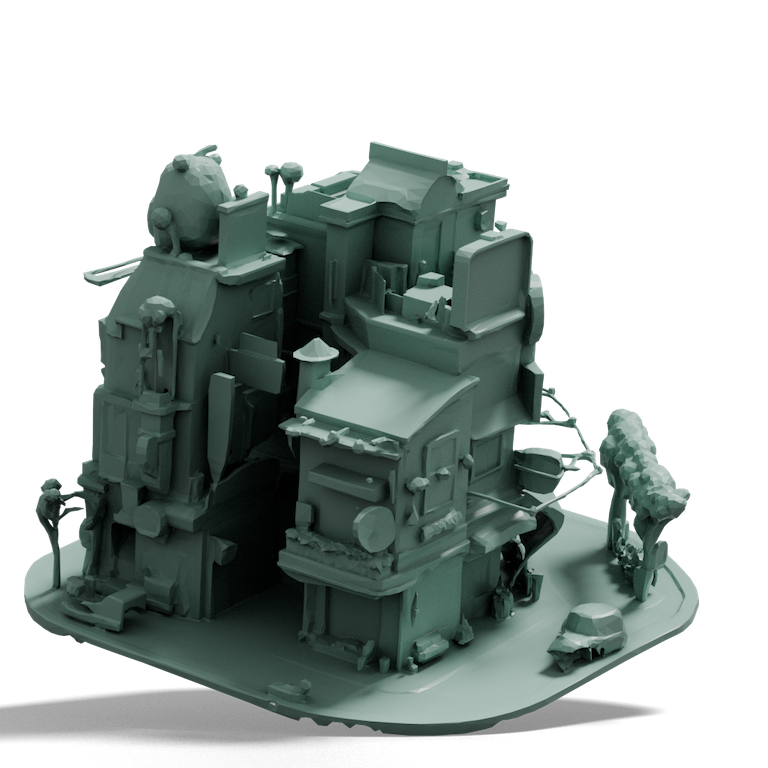} &
        \includegraphics[width=0.16\linewidth]{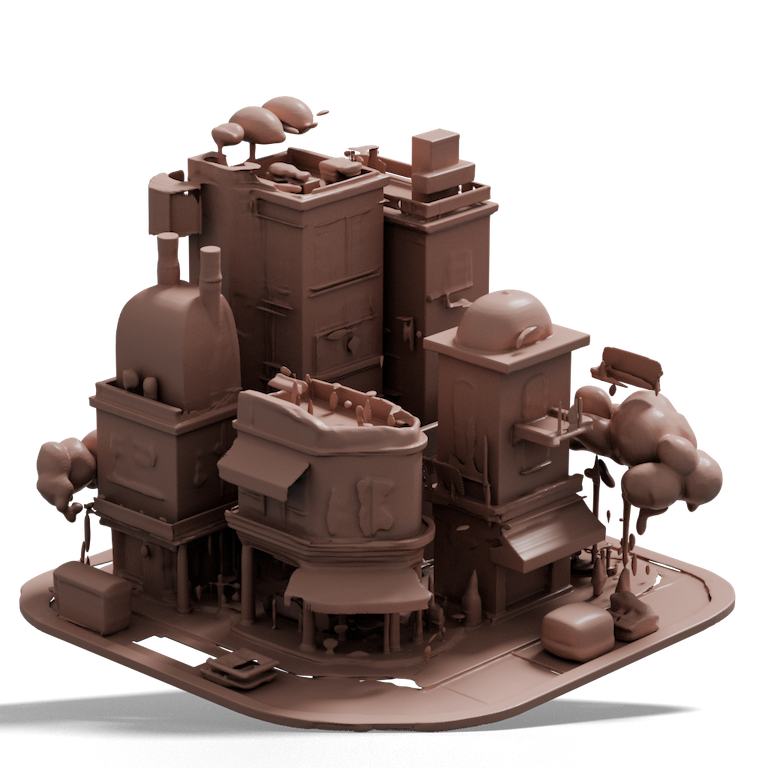} &
        \includegraphics[width=0.16\linewidth]{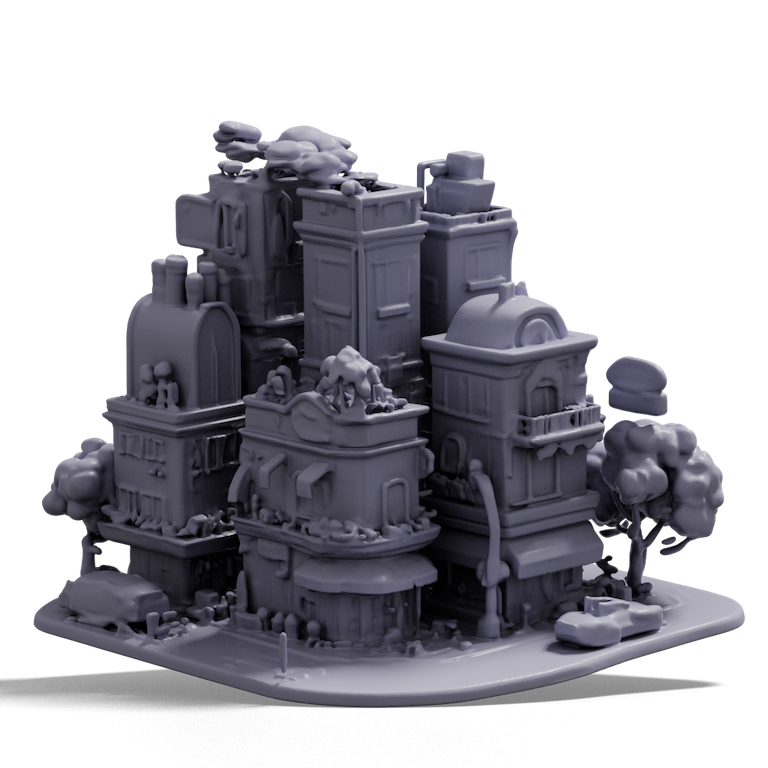} &
        \includegraphics[width=0.16\linewidth]{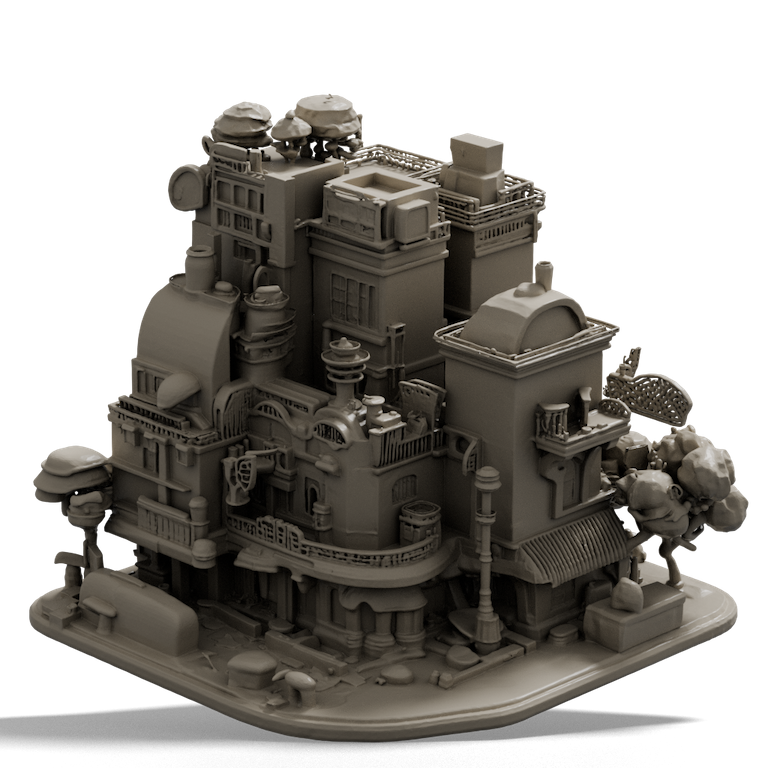} &
        \includegraphics[width=0.16\linewidth]{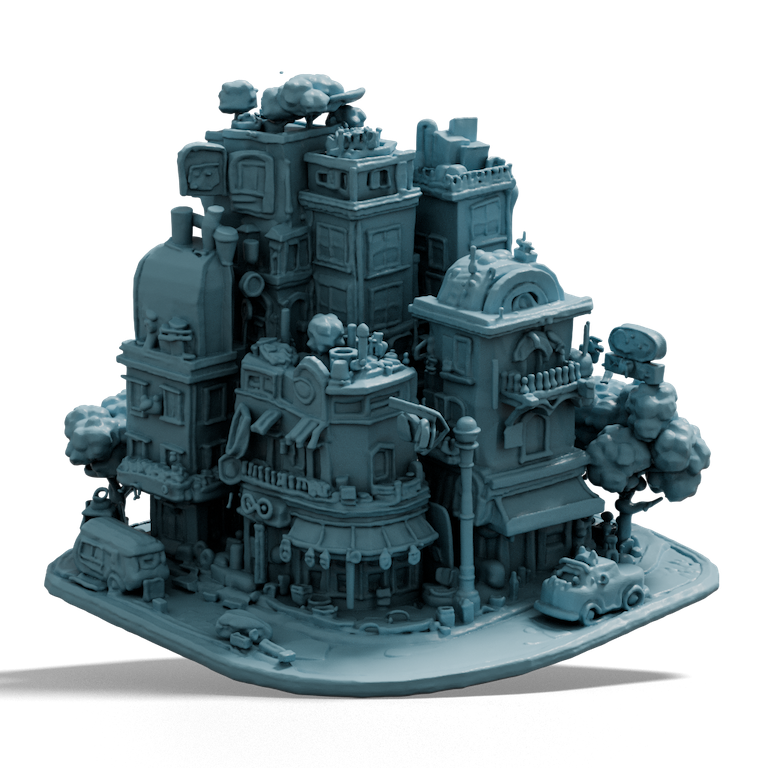} \\

        \includegraphics[width=0.16\linewidth]{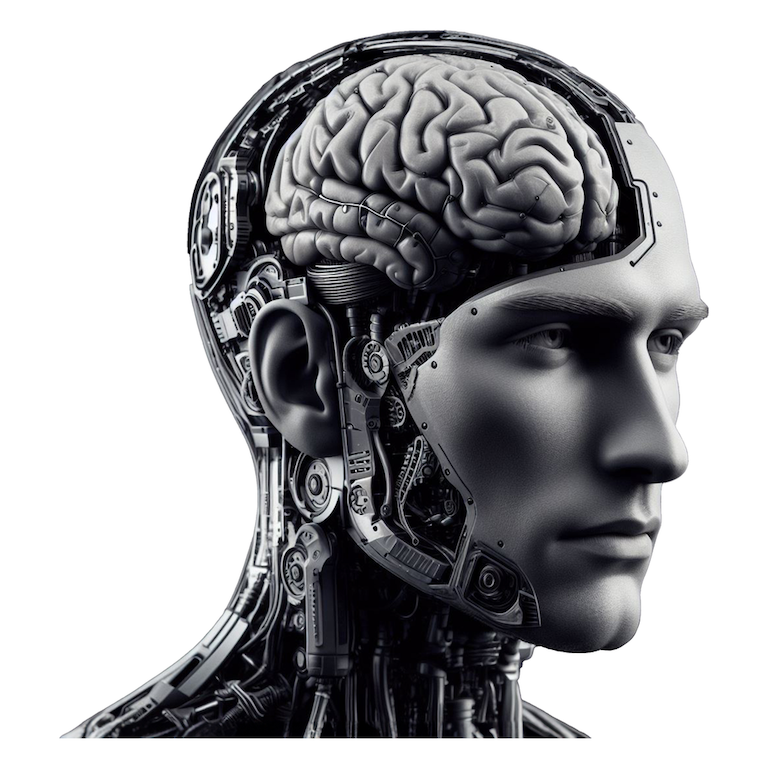} &
        \includegraphics[width=0.16\linewidth]{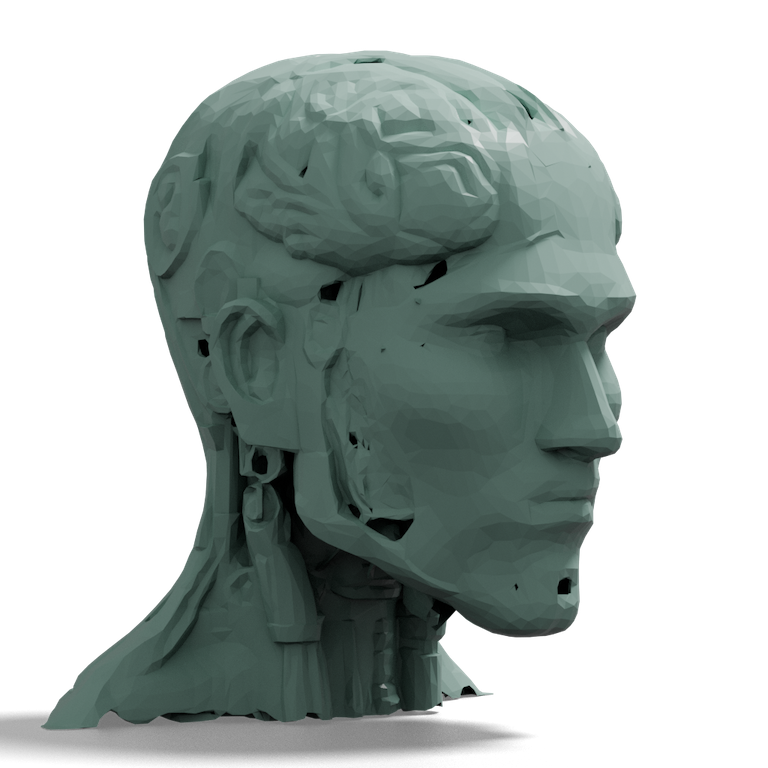} &
        \includegraphics[width=0.16\linewidth]{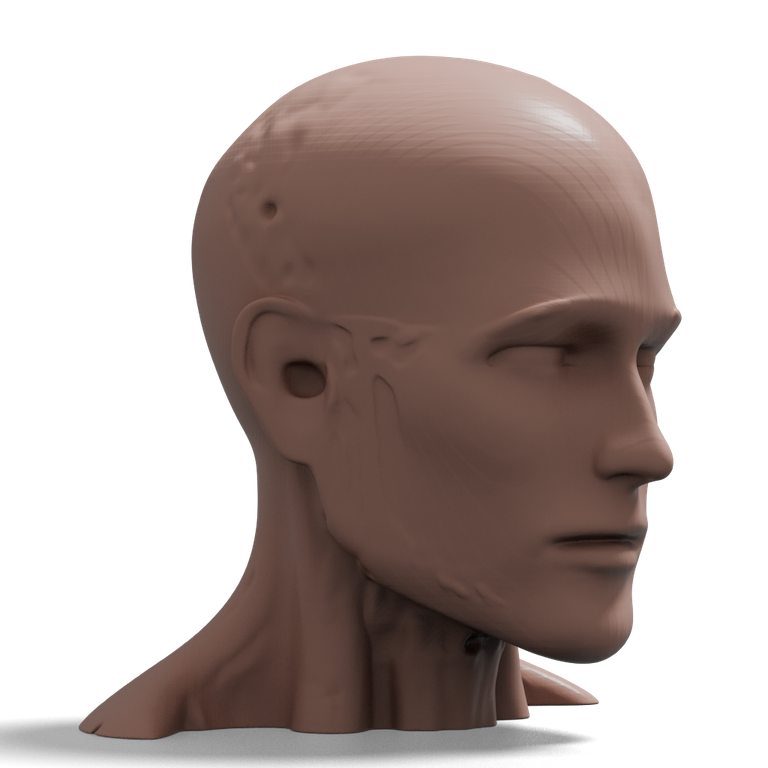} &
        \includegraphics[width=0.16\linewidth]{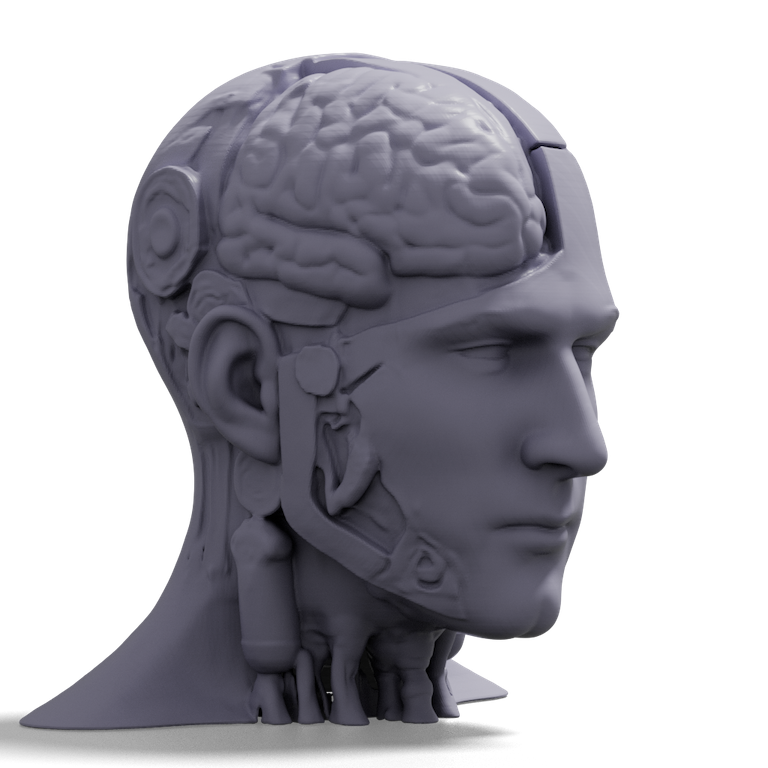} &
        \includegraphics[width=0.16\linewidth]{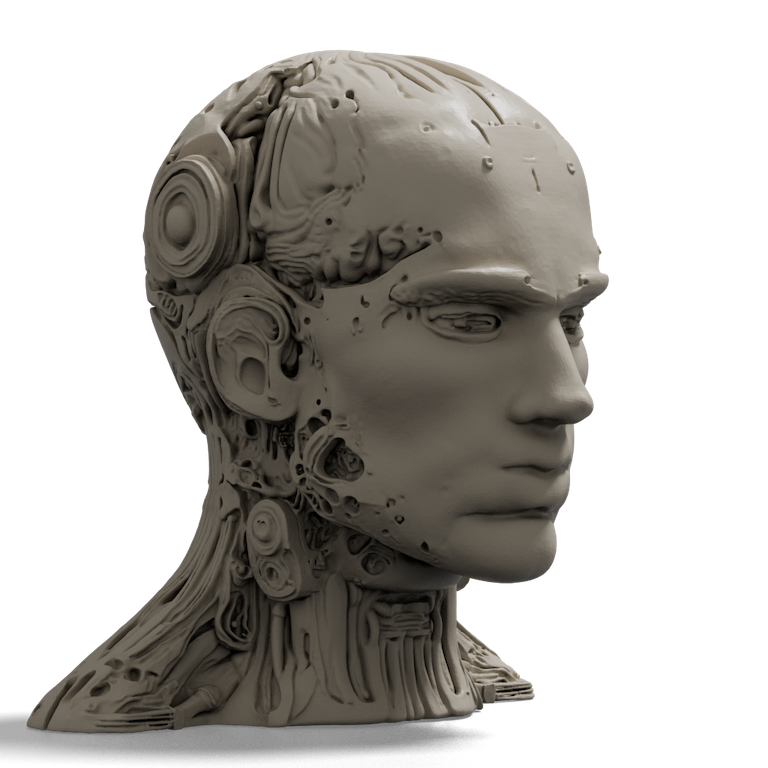} &
        \includegraphics[width=0.16\linewidth]{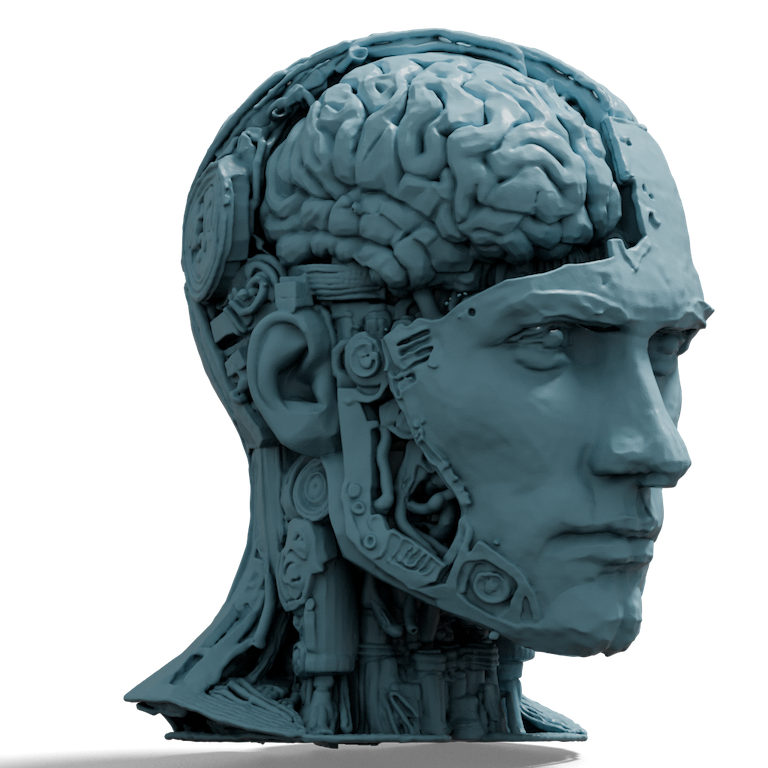} \\

        \includegraphics[width=0.16\linewidth]{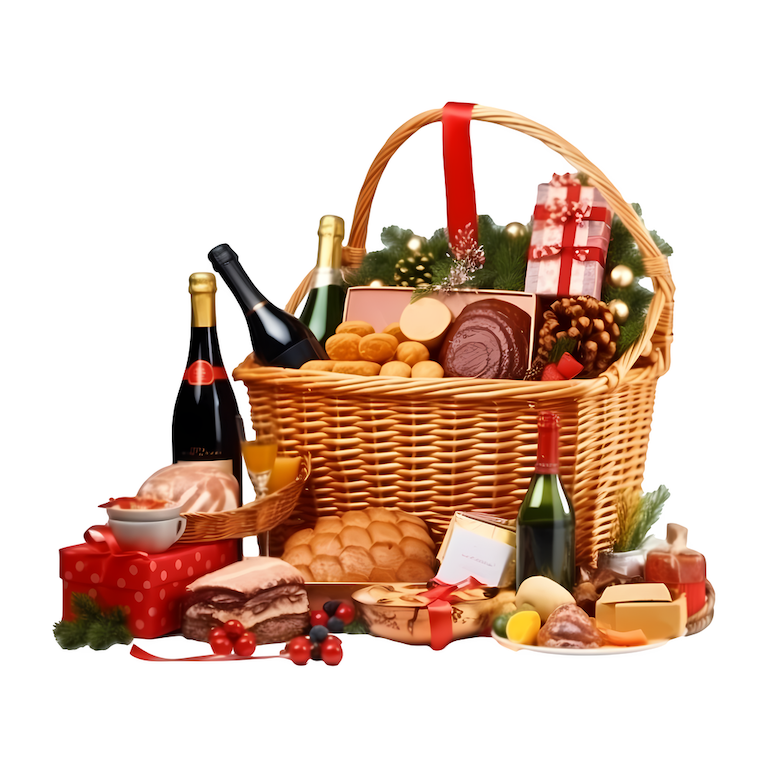} &
        \includegraphics[width=0.16\linewidth]{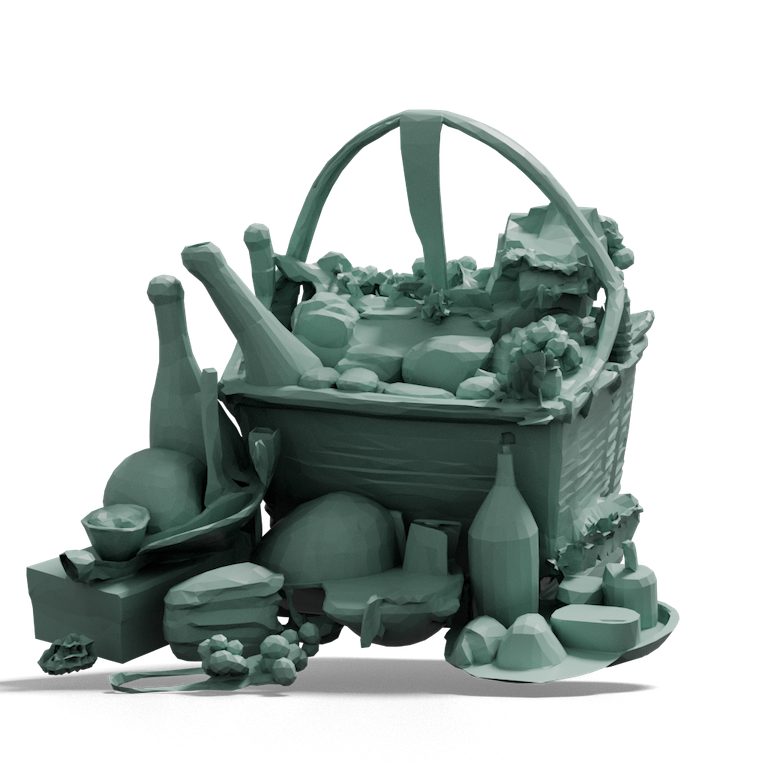} &
        \includegraphics[width=0.16\linewidth]{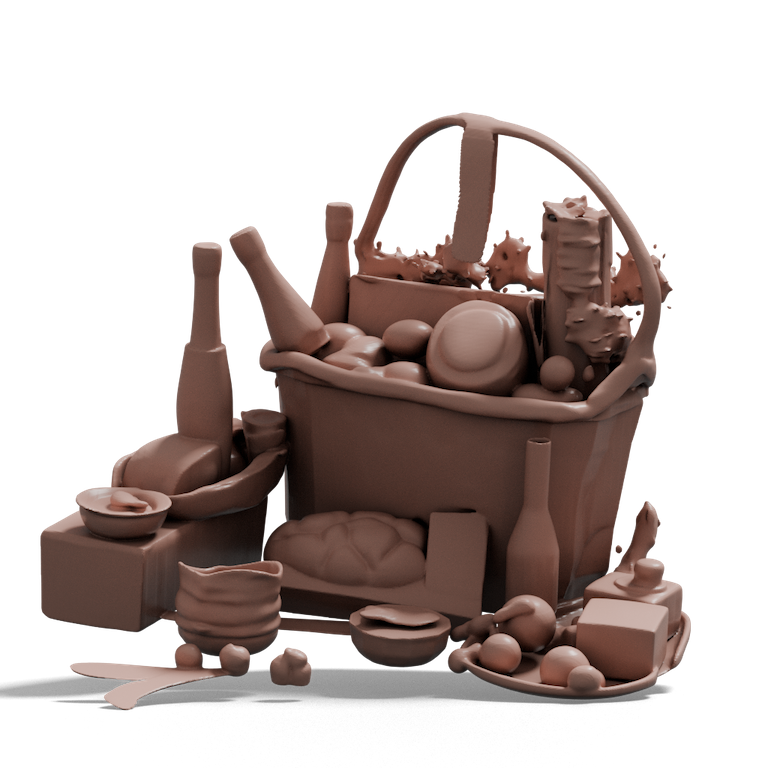} &
        \includegraphics[width=0.16\linewidth]{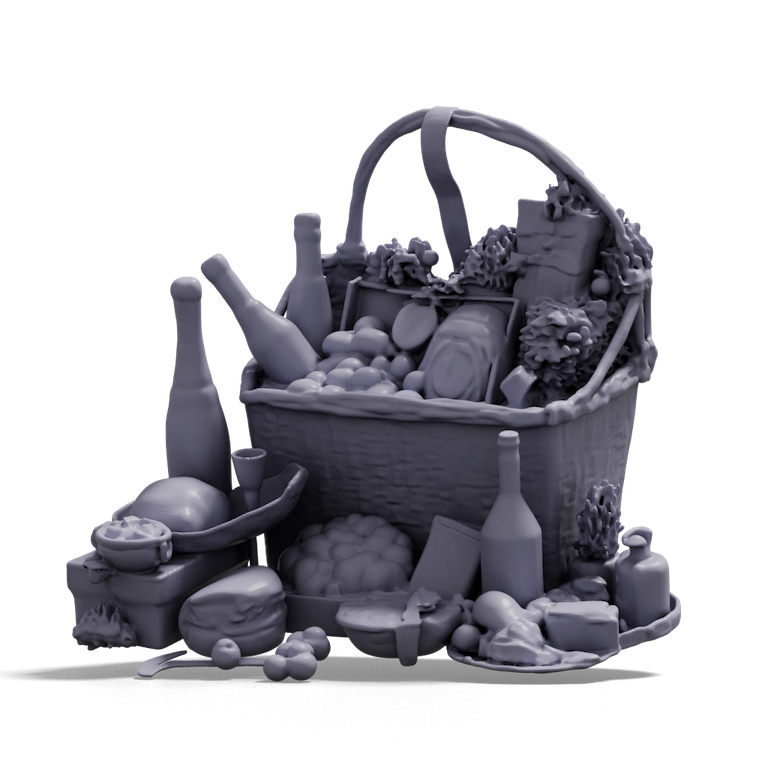} &
        \includegraphics[width=0.16\linewidth]{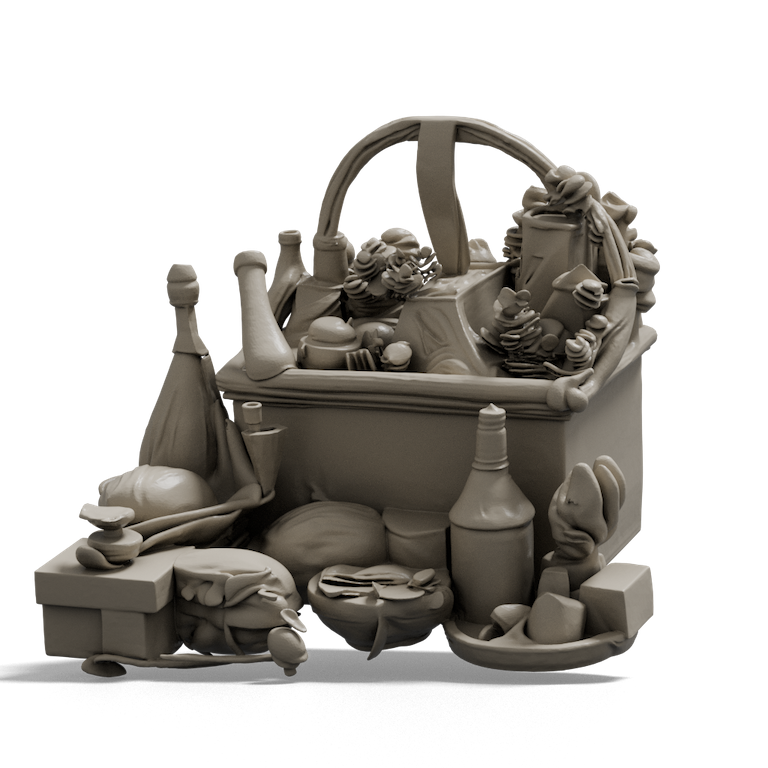} &
        \includegraphics[width=0.16\linewidth]{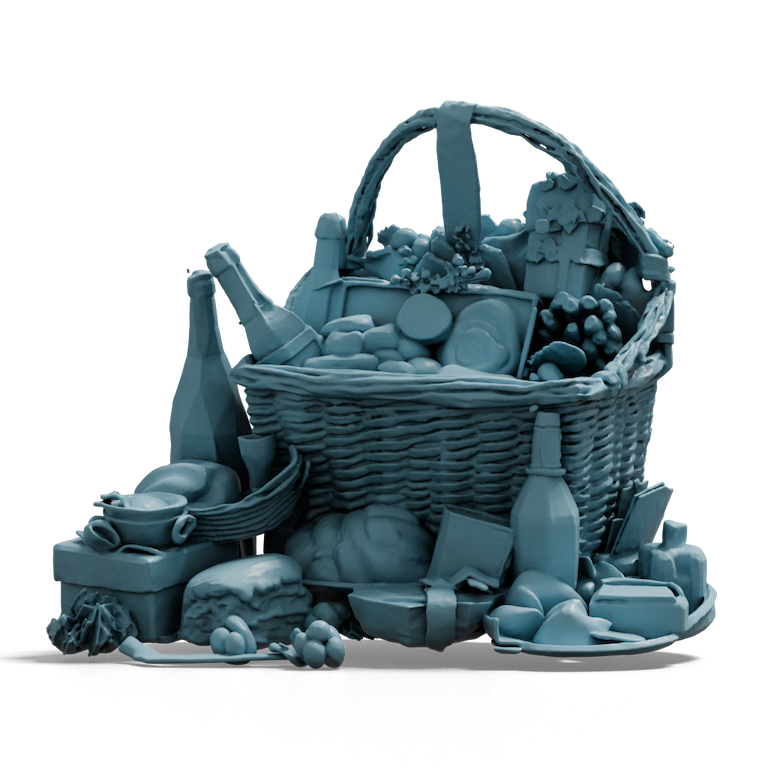} \\

        \includegraphics[width=0.16\linewidth]{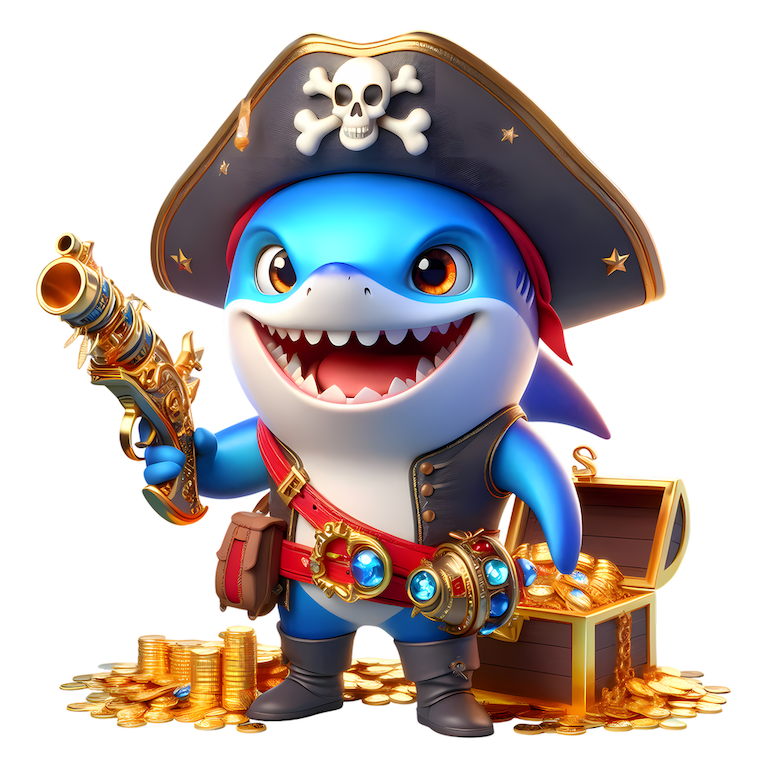} &
        \includegraphics[width=0.16\linewidth]{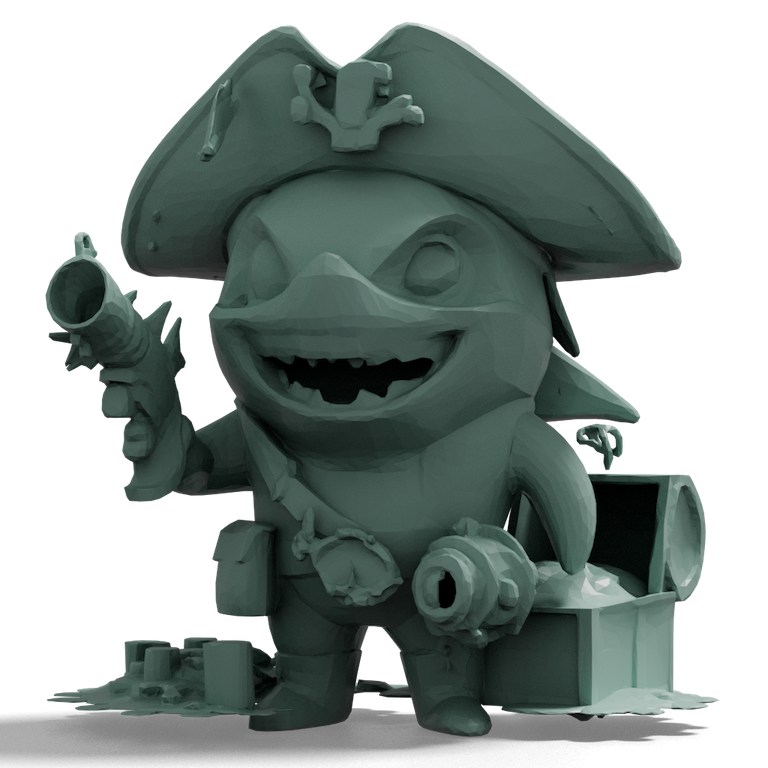} &
        \includegraphics[width=0.16\linewidth]{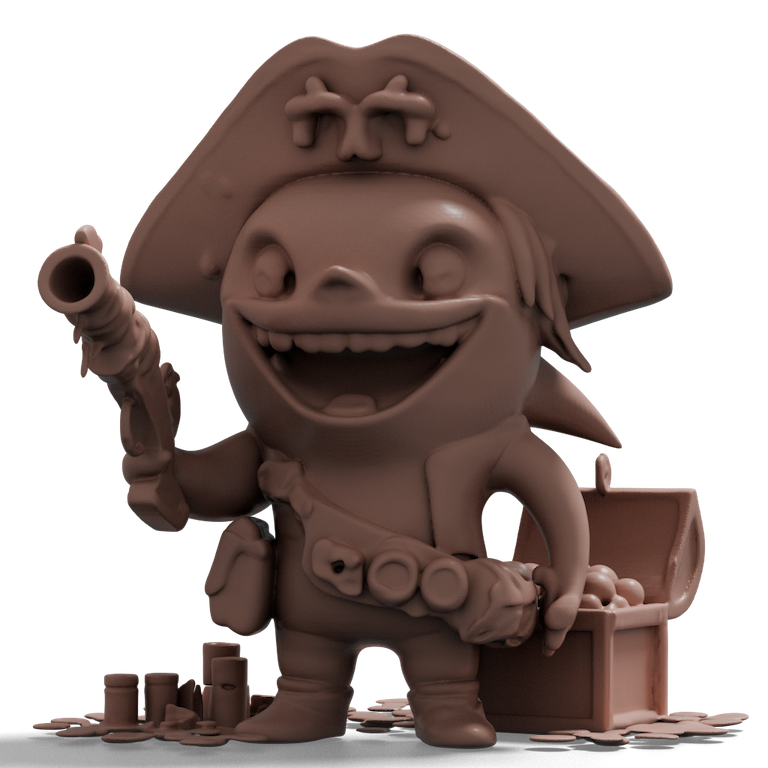} &
        \includegraphics[width=0.16\linewidth]{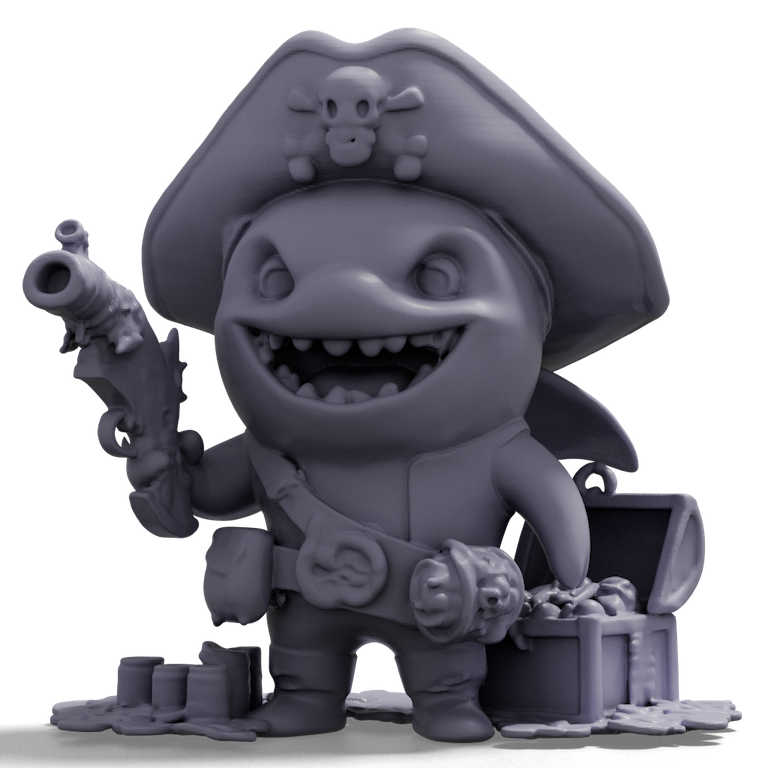} &
        \includegraphics[width=0.16\linewidth]{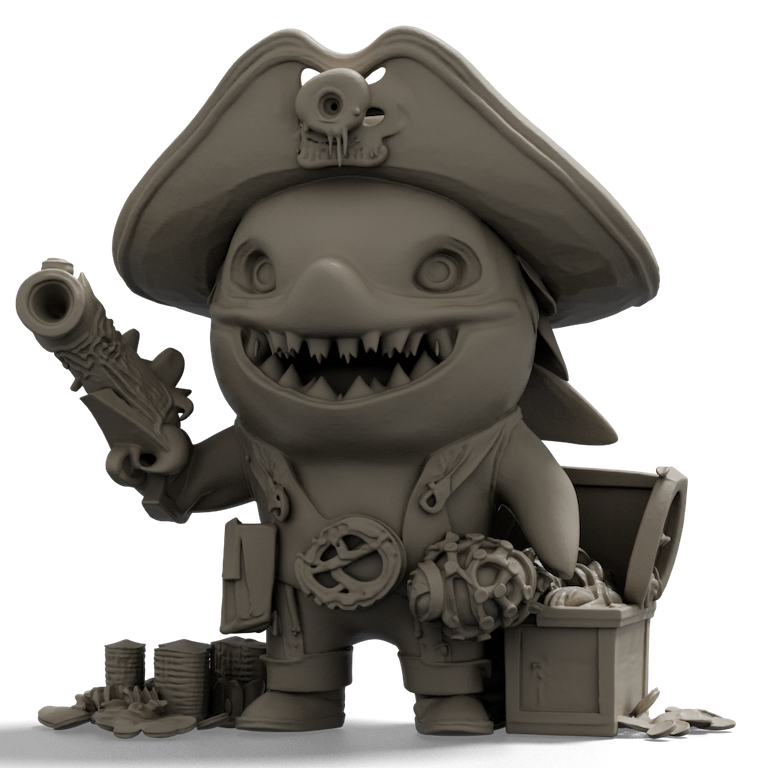} &
        \includegraphics[width=0.16\linewidth]{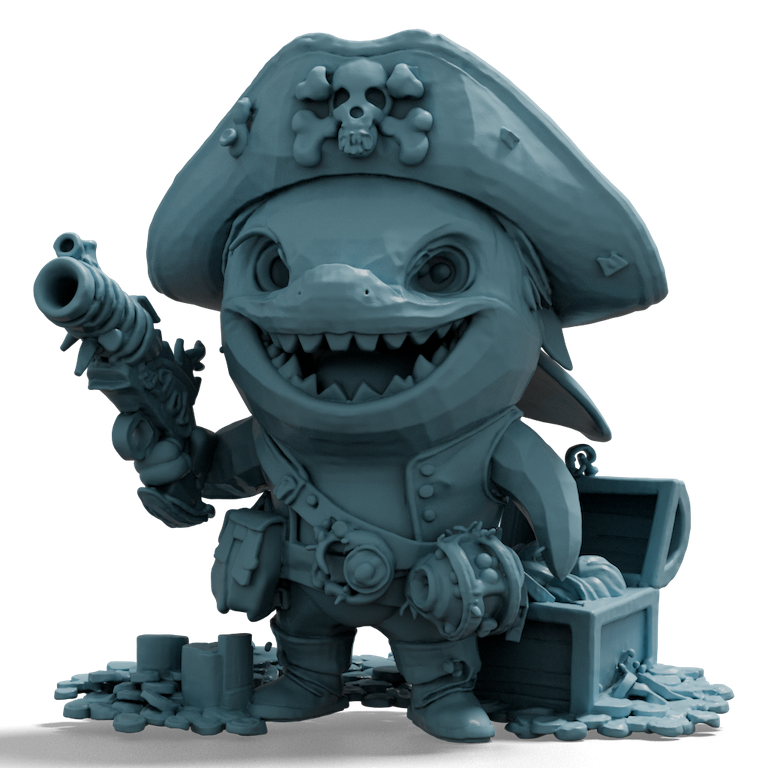} \\

        \includegraphics[width=0.16\linewidth]{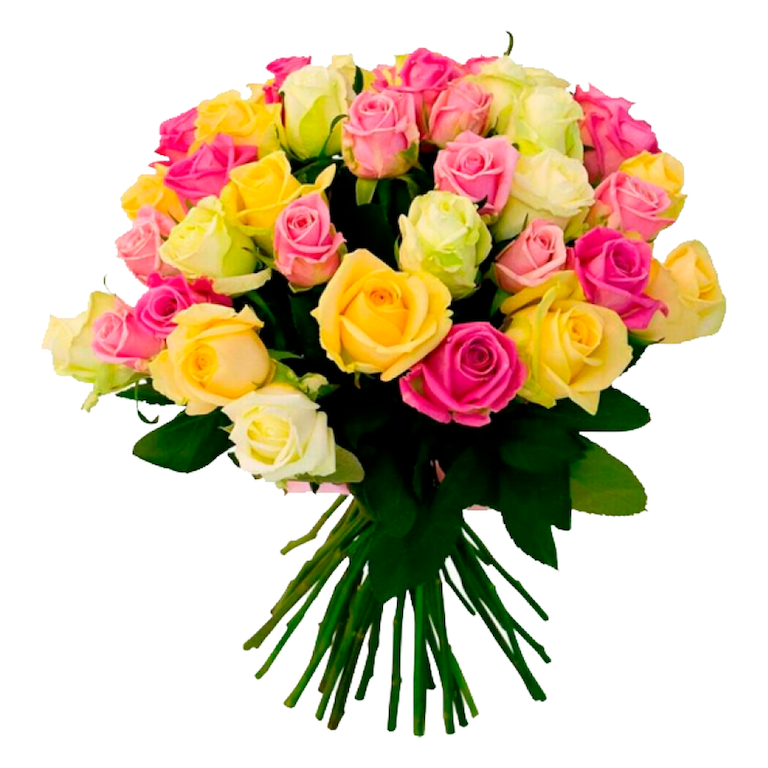} &
        \includegraphics[width=0.16\linewidth]{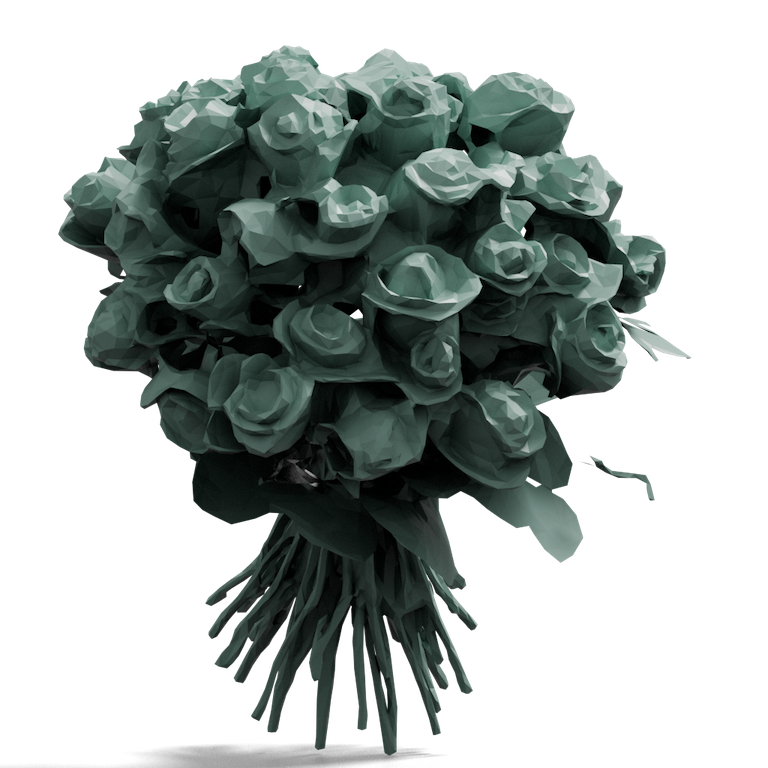} &
        \includegraphics[width=0.16\linewidth]{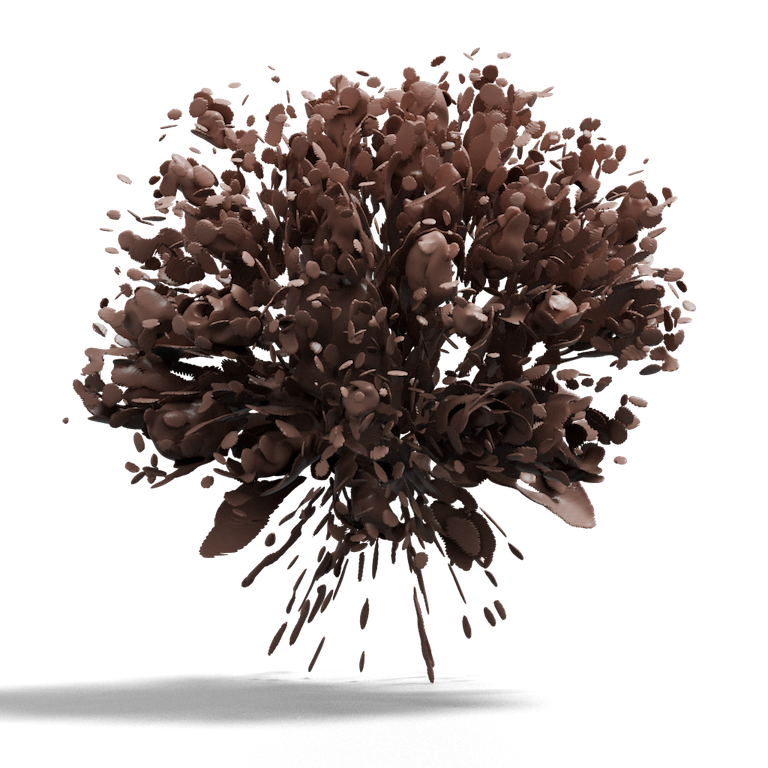} &
        \includegraphics[width=0.16\linewidth]{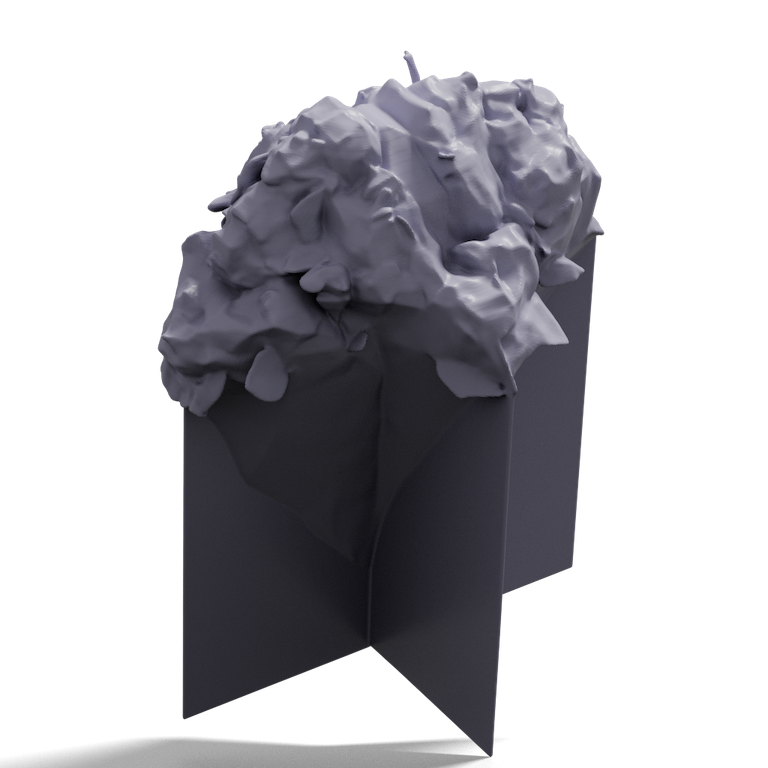} &
        \includegraphics[width=0.16\linewidth]{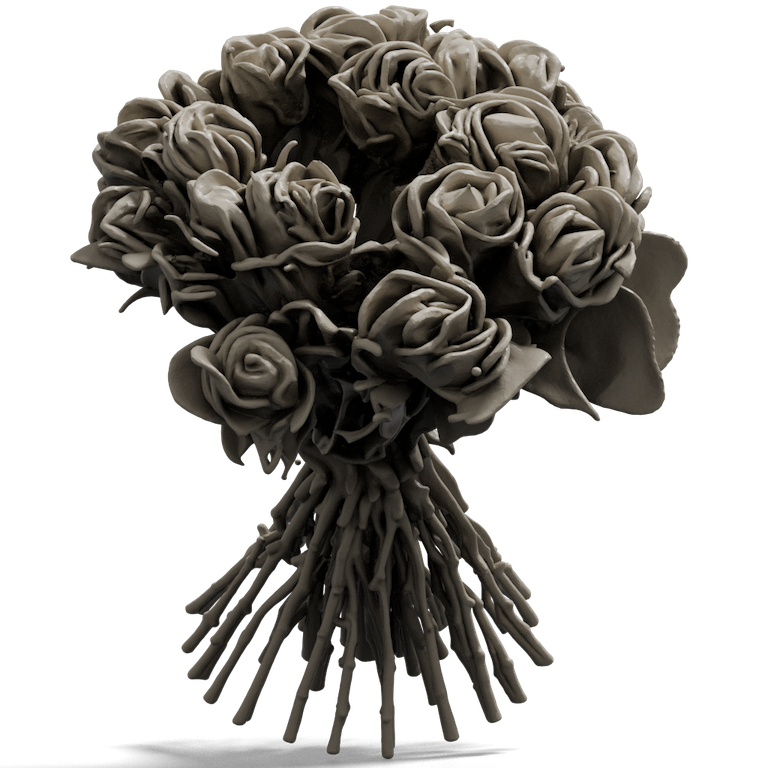} &
        \includegraphics[width=0.16\linewidth]{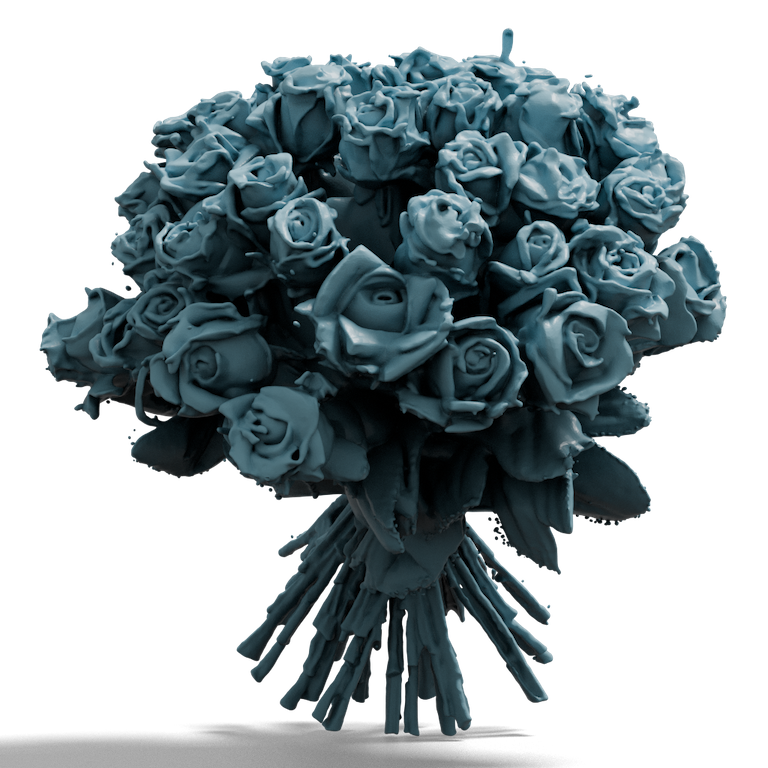} \\

        \includegraphics[width=0.16\linewidth]{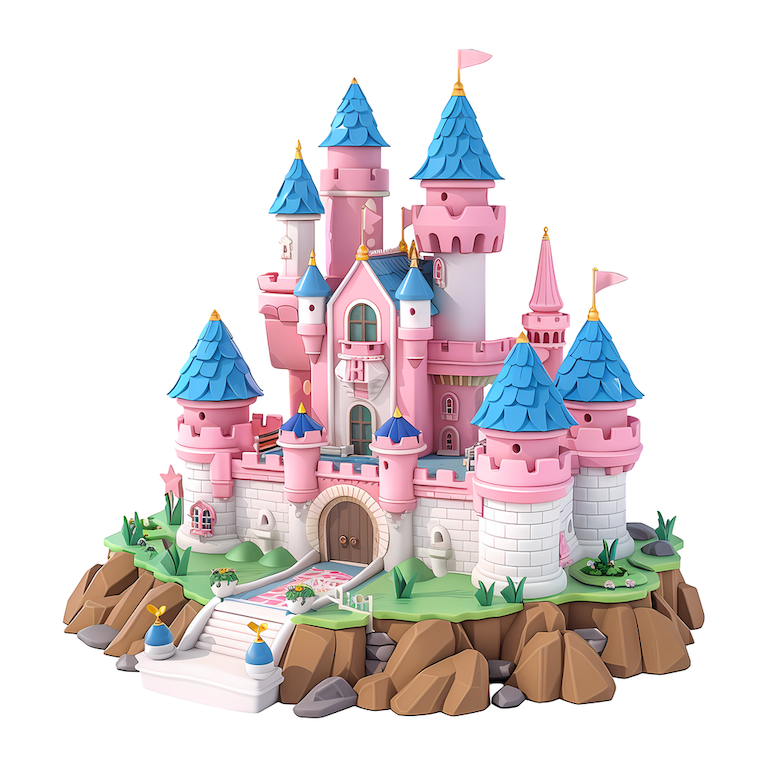} &
        \includegraphics[width=0.16\linewidth]{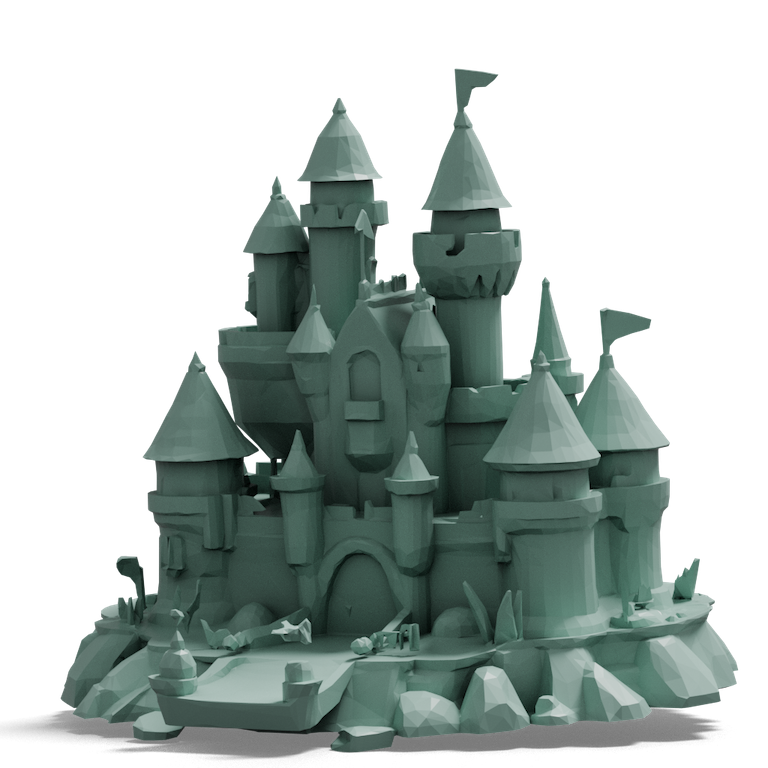} &
        \includegraphics[width=0.16\linewidth]{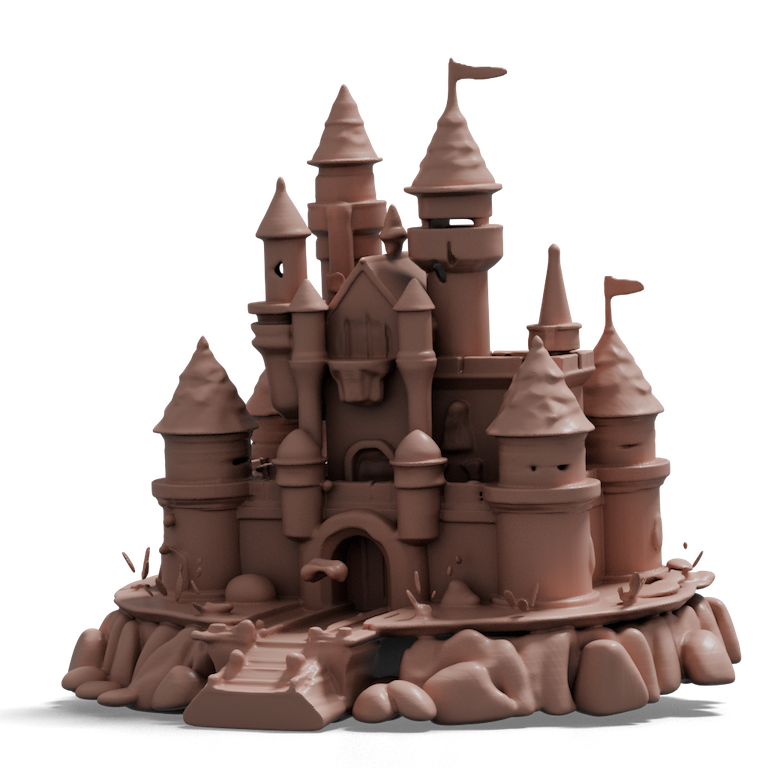} &
        \includegraphics[width=0.16\linewidth]{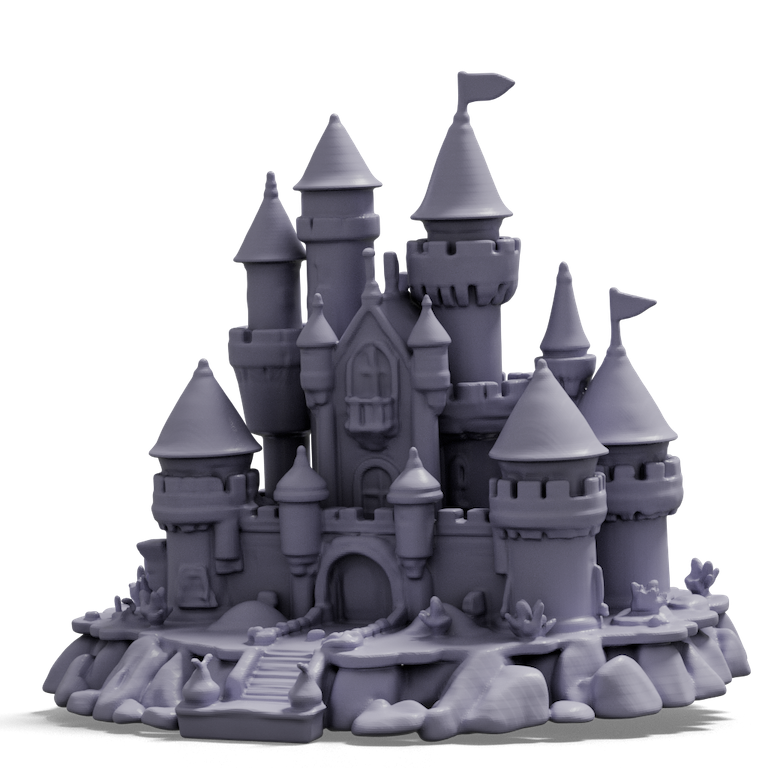} &
        \includegraphics[width=0.16\linewidth]{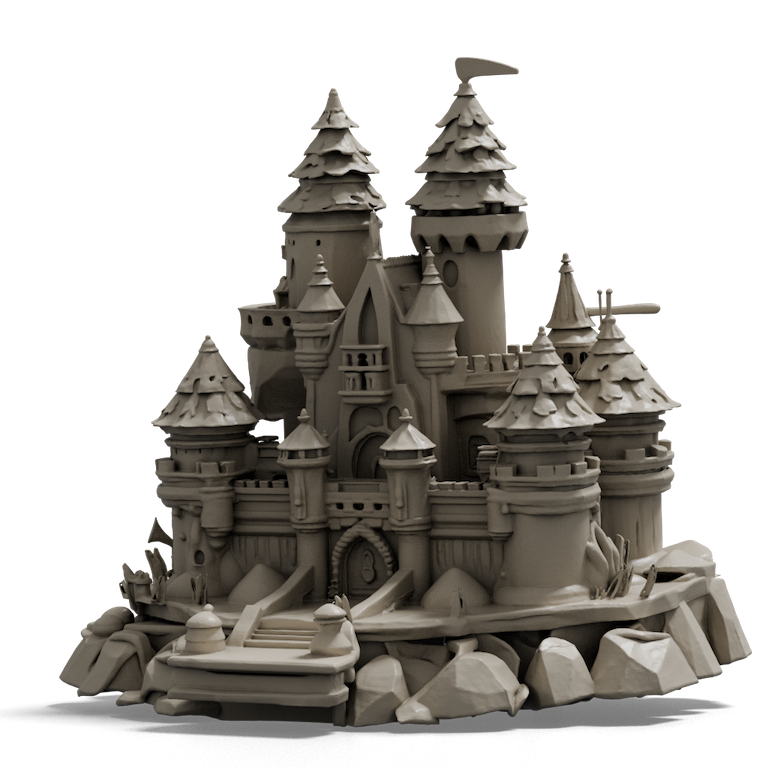} &
        \includegraphics[width=0.16\linewidth]{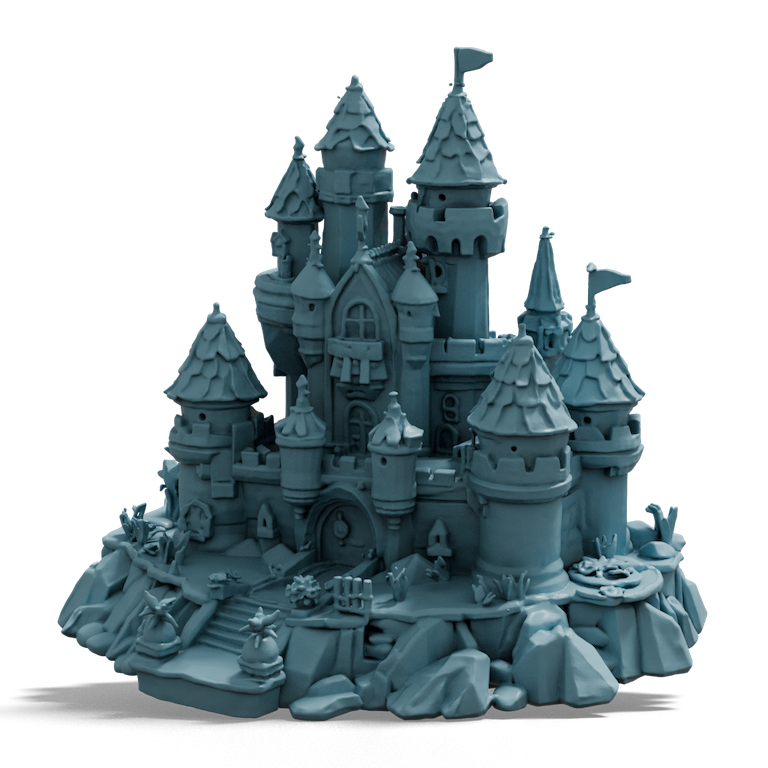} \\

        Input & TRELLIS & TripoSG & Hunyuan3D-2.1 & Direct3D-S2 & Pixal3D  \\
    \end{tabular}

    \caption{Qualitative comparison of single-view 3D generation on in-the-wild images.}
    \label{fig_only:singleiview}
\end{figure*}

% \begin{figure*}[!p]
% % \begin{figure*}[!t]
%     \vskip -0.2in
%    \newcommand{\multiviewrow}[1]{
%     \includegraphics[width=0.23\linewidth]{figures/images/multiview_figure_image/#1_images_4_view_grid.png} &
%     \includegraphics[width=0.23\linewidth]{figures/images/multiview_figure_image/#1_vggt_4_view_grid.png} &
%     \includegraphics[width=0.23\linewidth]{figures/images/multiview_figure_image/#1_trellis_4_view_grid.png} &
%     \includegraphics[width=0.23\linewidth]{figures/images/multiview_figure_image/#1_pa3d_4_view_grid.png} 
 
%     \\
% }
%     \centering
%     \small
%     \setlength{\tabcolsep}{0pt}
%     \begin{tabular}{cccc}

%         \centering
        
%         \multiviewrow{a65ef}
%         \multiviewrow{a003d}
%         \multiviewrow{29355}
%         \multiviewrow{9df0e}
%         \multiviewrow{3d693}

%        Input & VGGT & TRELLIS & Pixal3D  \\
%     \end{tabular}
%     \caption{Qualitative comparison of multi-view 3D generation on Toys4K. }
%     \label{fig:multiview}
% \end{figure*}
\begin{figure*}[!p]
% \begin{figure*}[!t]
    \vskip -0.2in
    \centering
    \small
    \setlength{\tabcolsep}{0pt}

    \begin{tabular}{cccc}

        \includegraphics[width=0.23\linewidth]{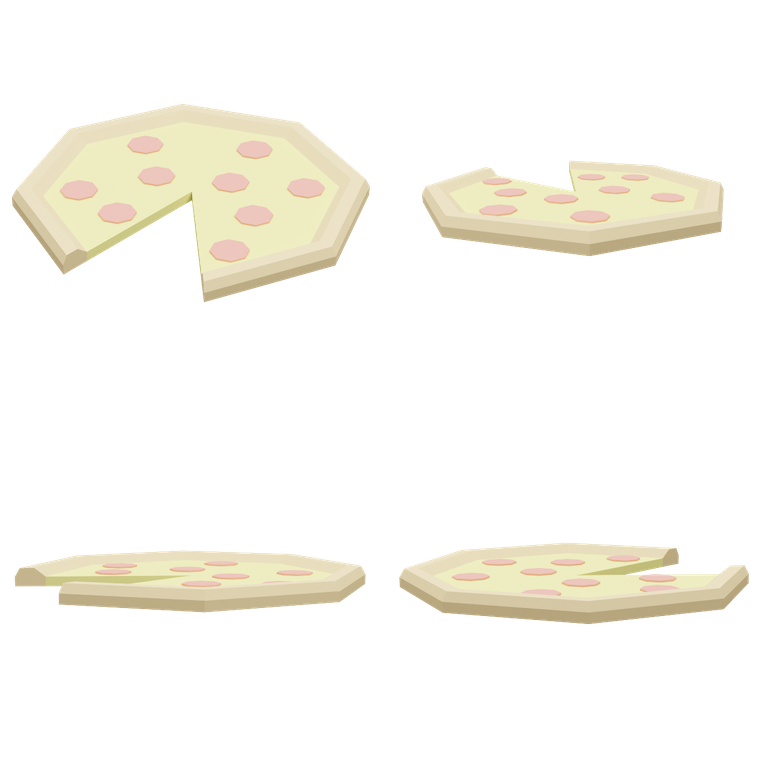} &
        \includegraphics[width=0.23\linewidth]{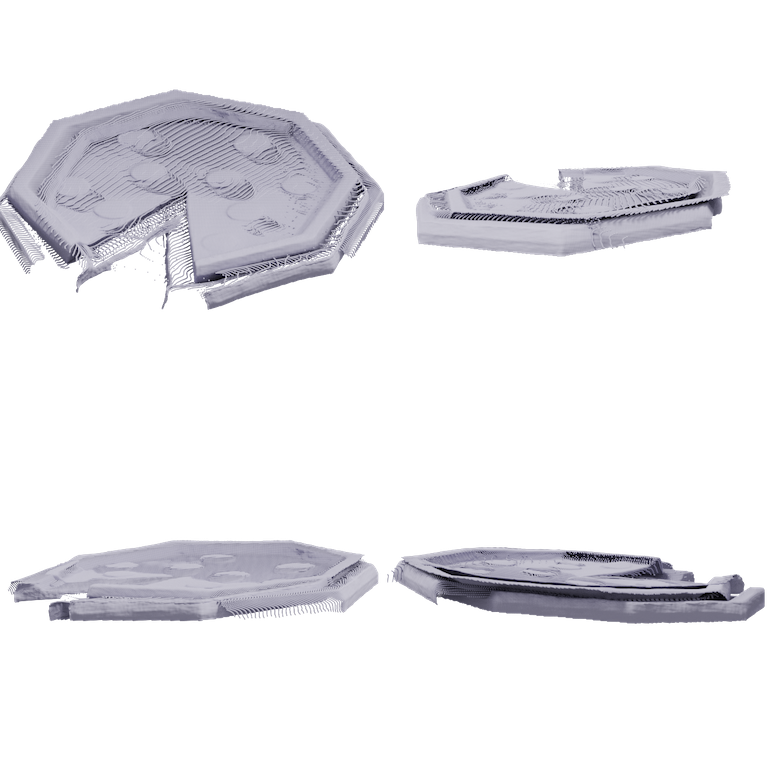} &
        \includegraphics[width=0.23\linewidth]{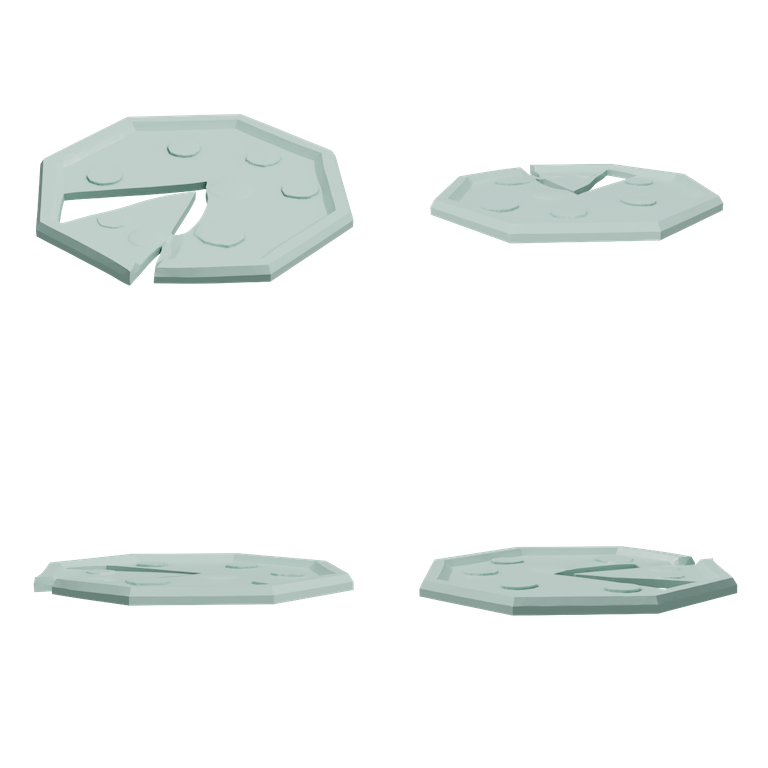} &
        \includegraphics[width=0.23\linewidth]{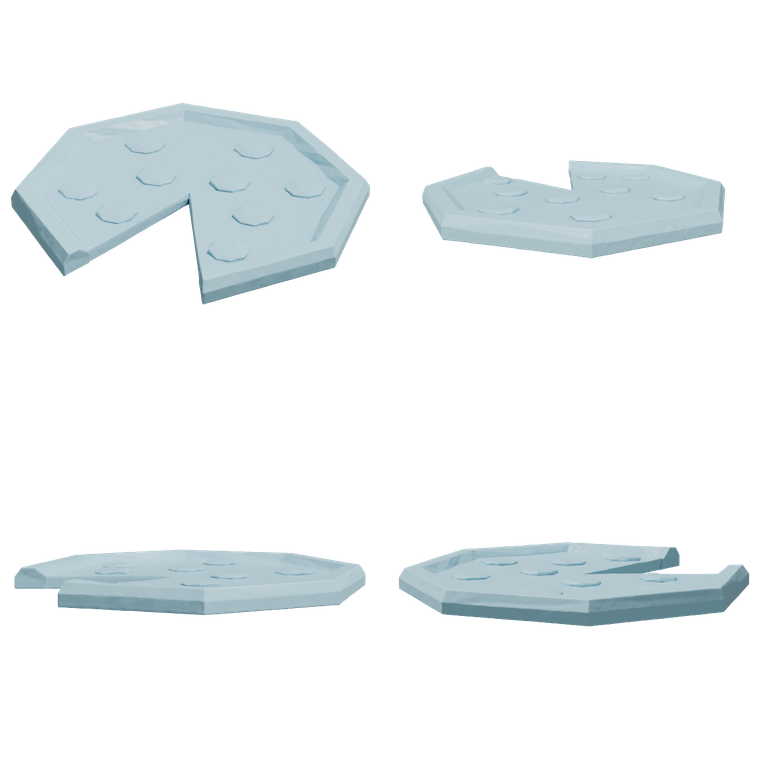} \\

        \includegraphics[width=0.23\linewidth]{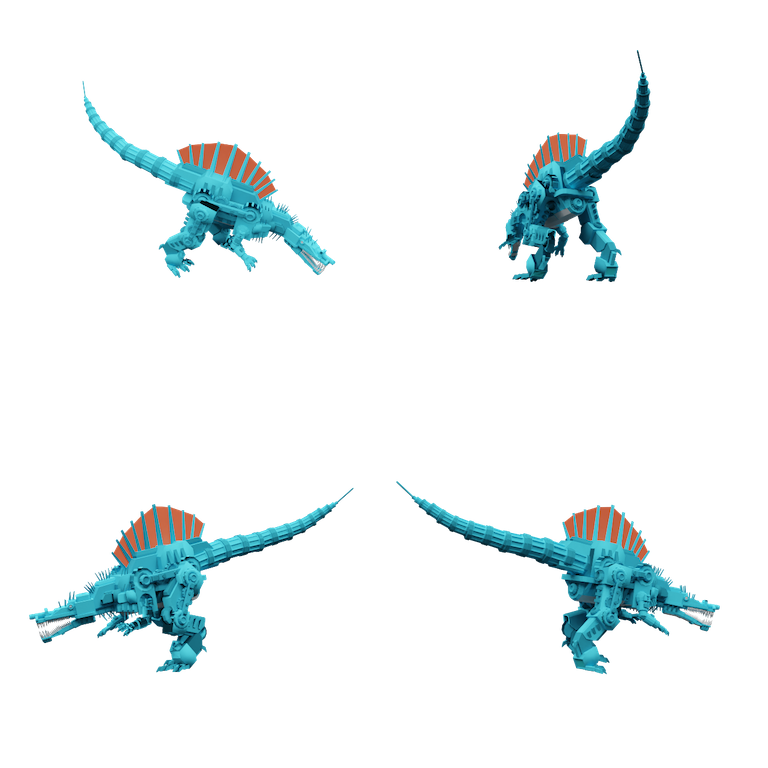} &
        \includegraphics[width=0.23\linewidth]{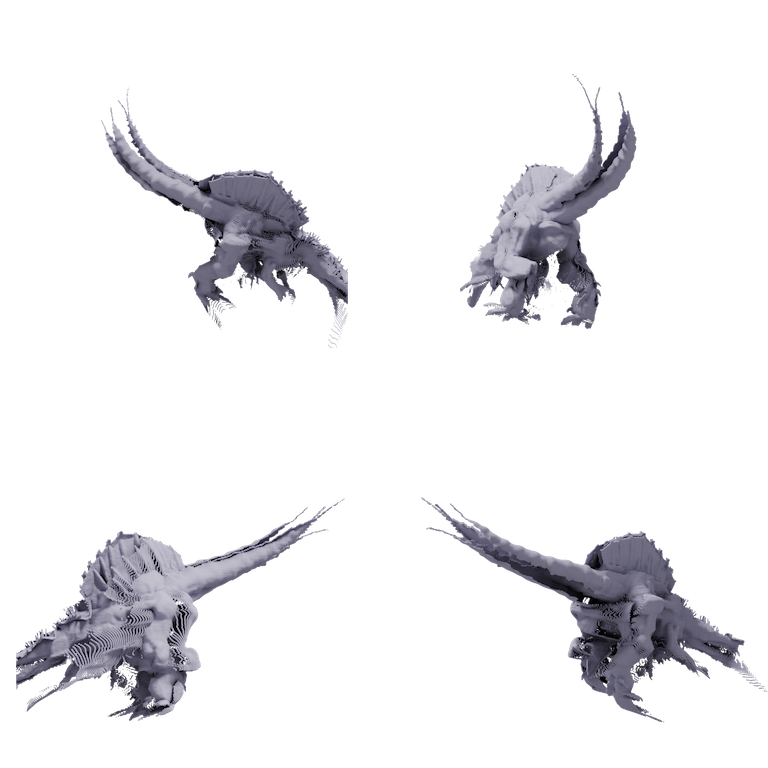} &
        \includegraphics[width=0.23\linewidth]{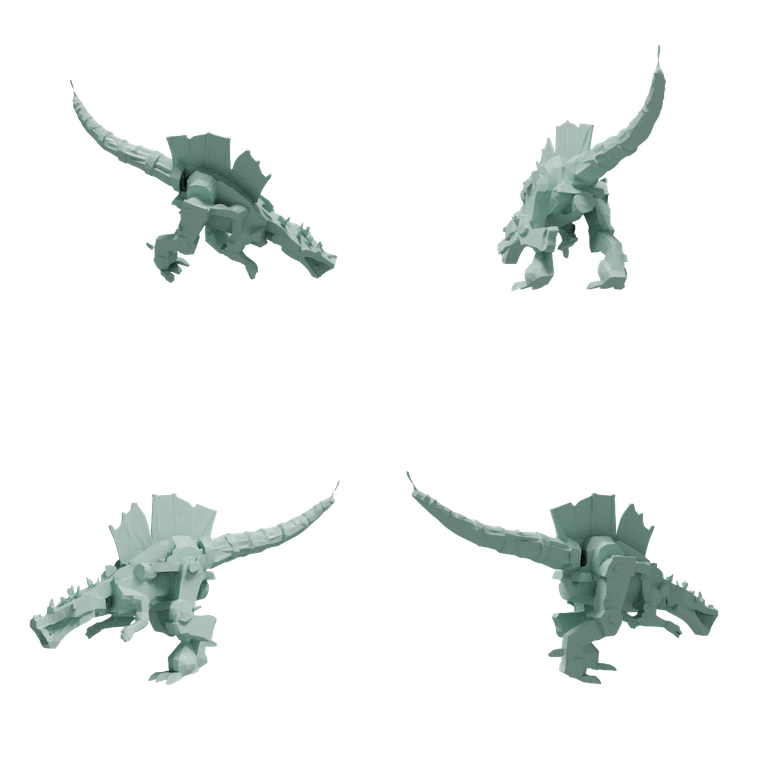} &
        \includegraphics[width=0.23\linewidth]{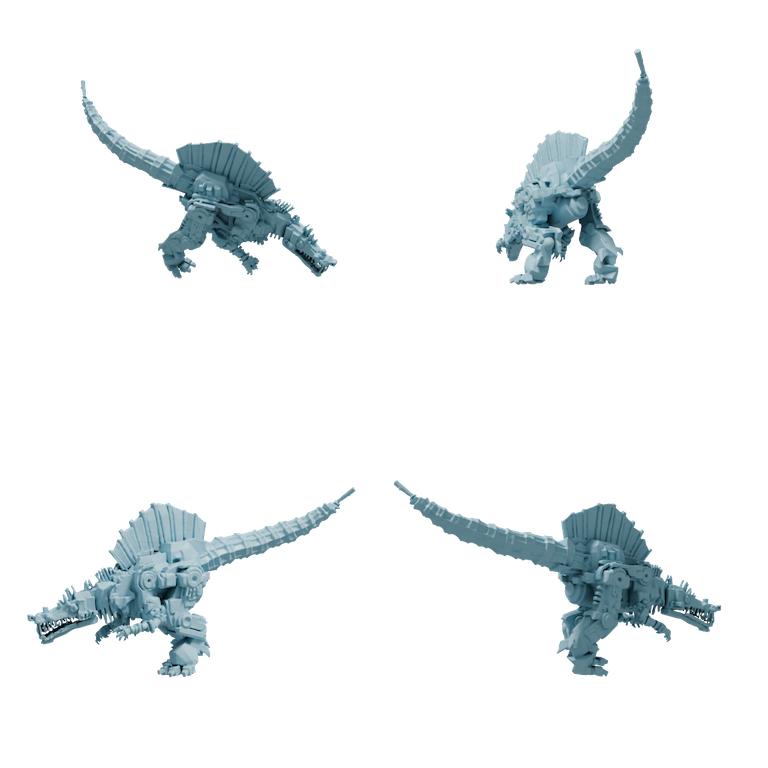} \\

        \includegraphics[width=0.23\linewidth]{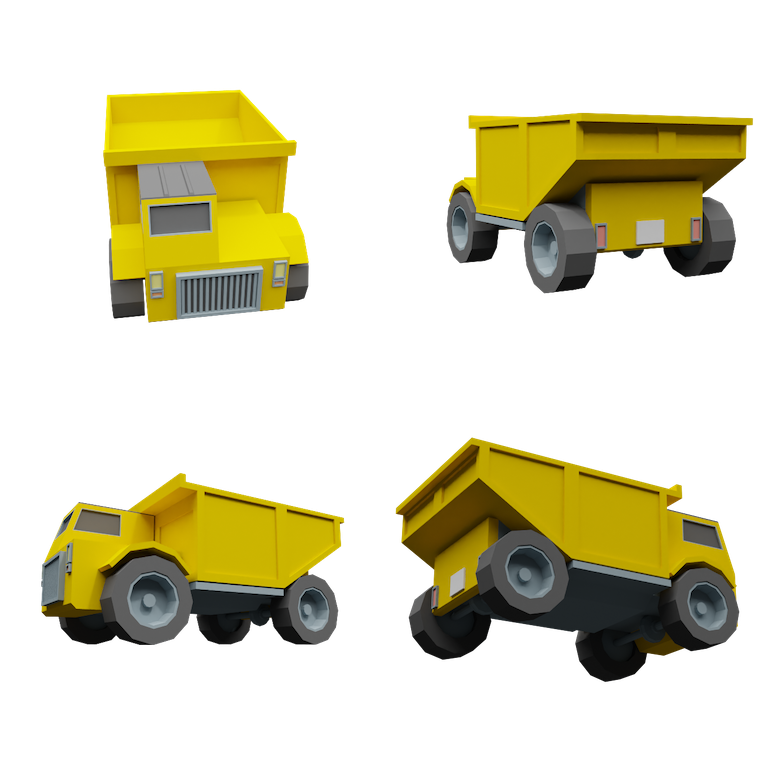} &
        \includegraphics[width=0.23\linewidth]{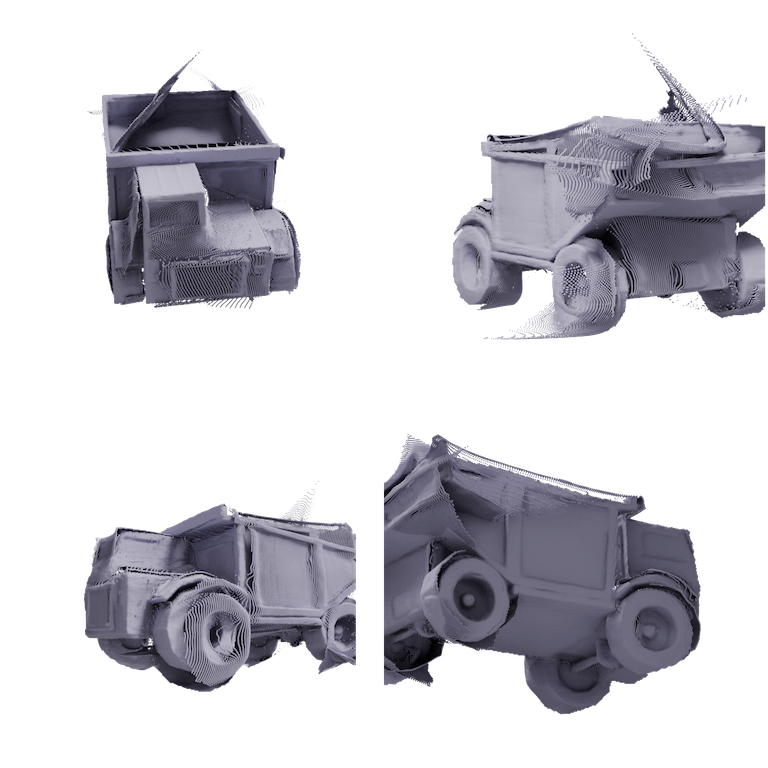} &
        \includegraphics[width=0.23\linewidth]{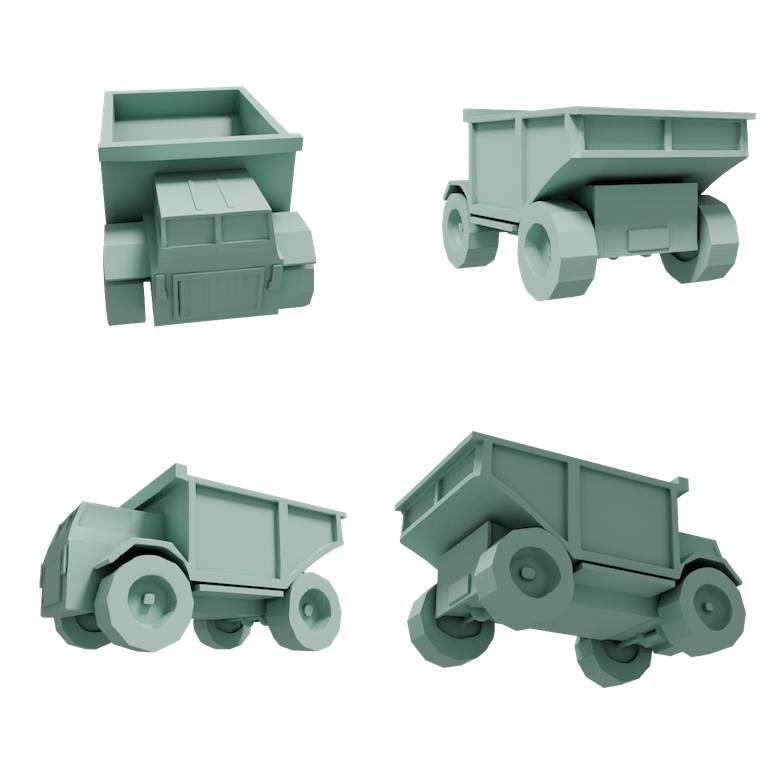} &
        \includegraphics[width=0.23\linewidth]{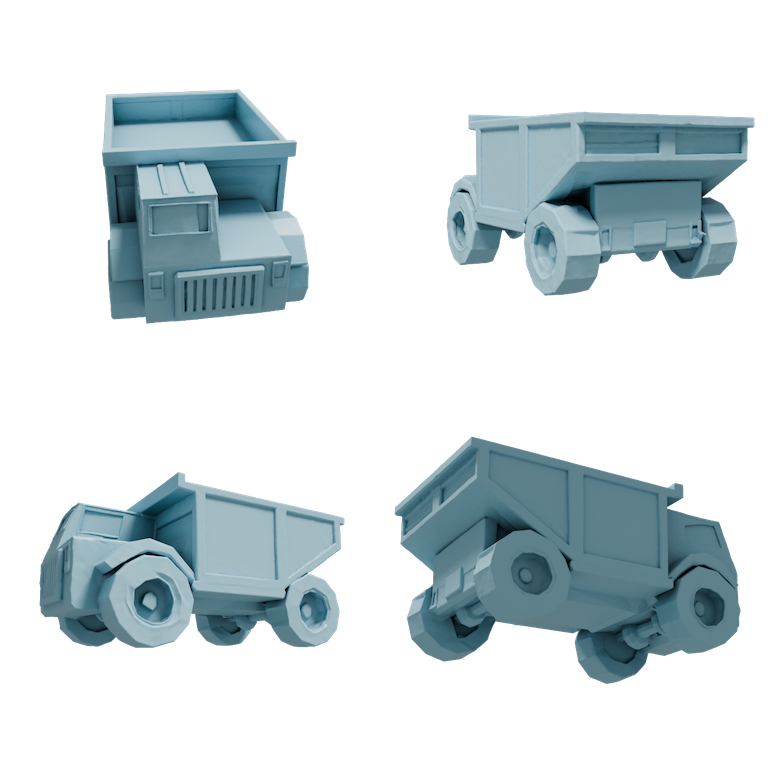} \\

        \includegraphics[width=0.23\linewidth]{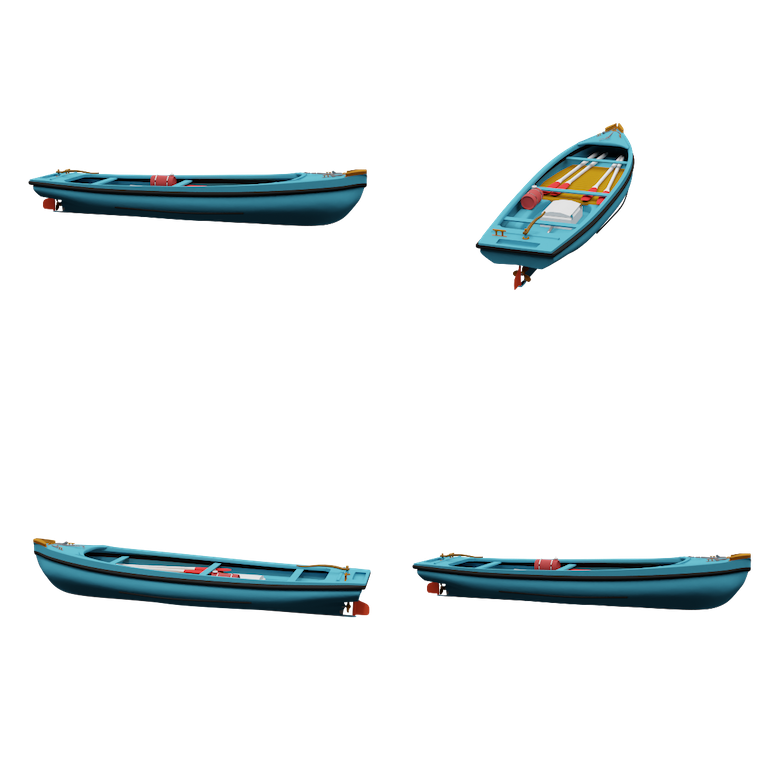} &
        \includegraphics[width=0.23\linewidth]{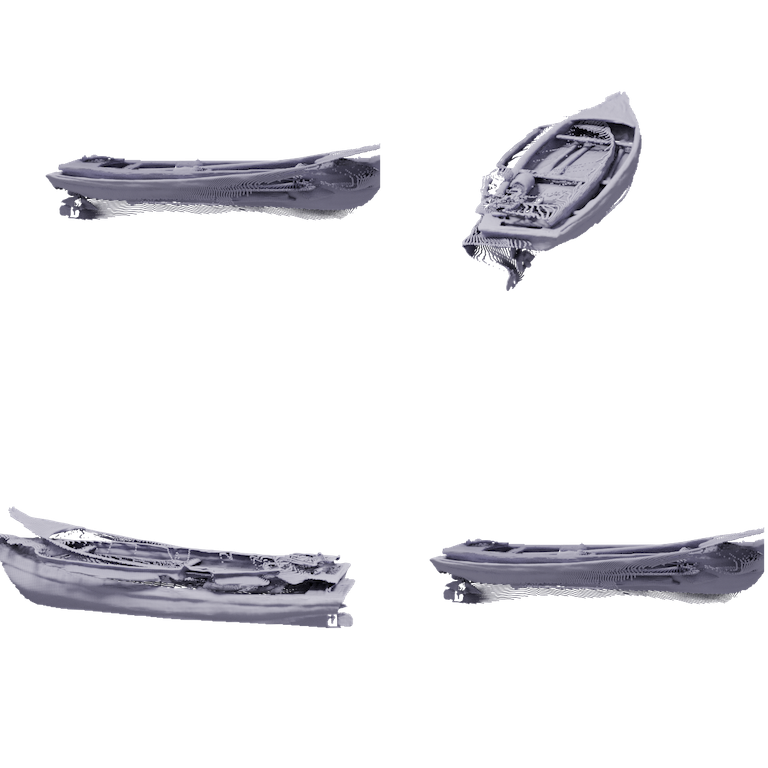} &
        \includegraphics[width=0.23\linewidth]{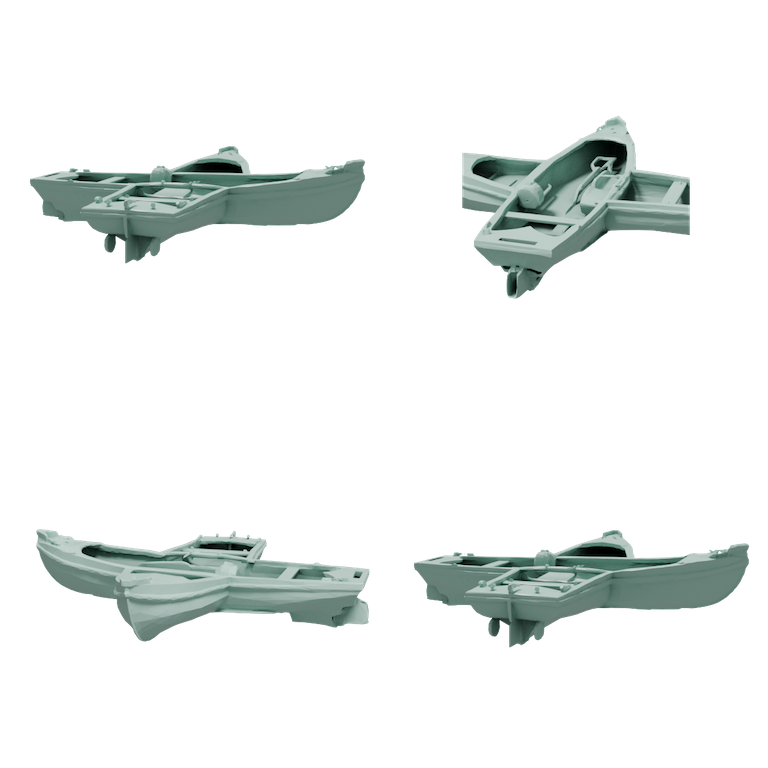} &
        \includegraphics[width=0.23\linewidth]{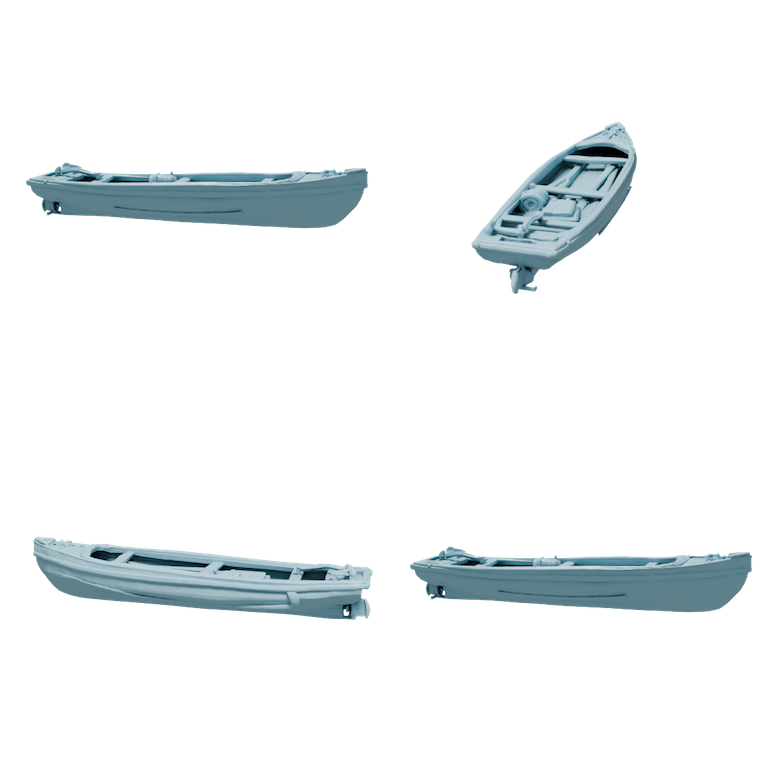} \\

        \includegraphics[width=0.23\linewidth]{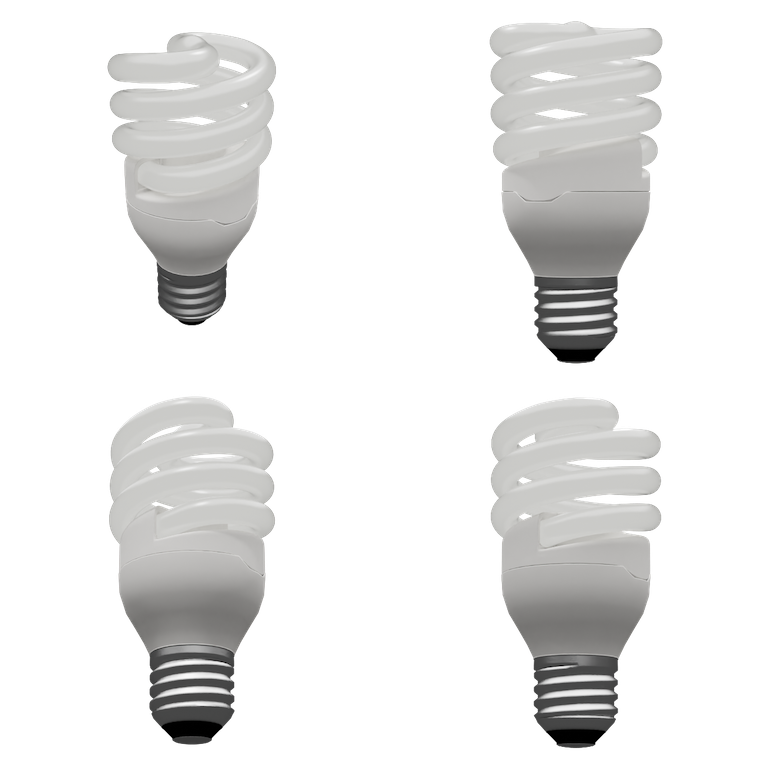} &
        \includegraphics[width=0.23\linewidth]{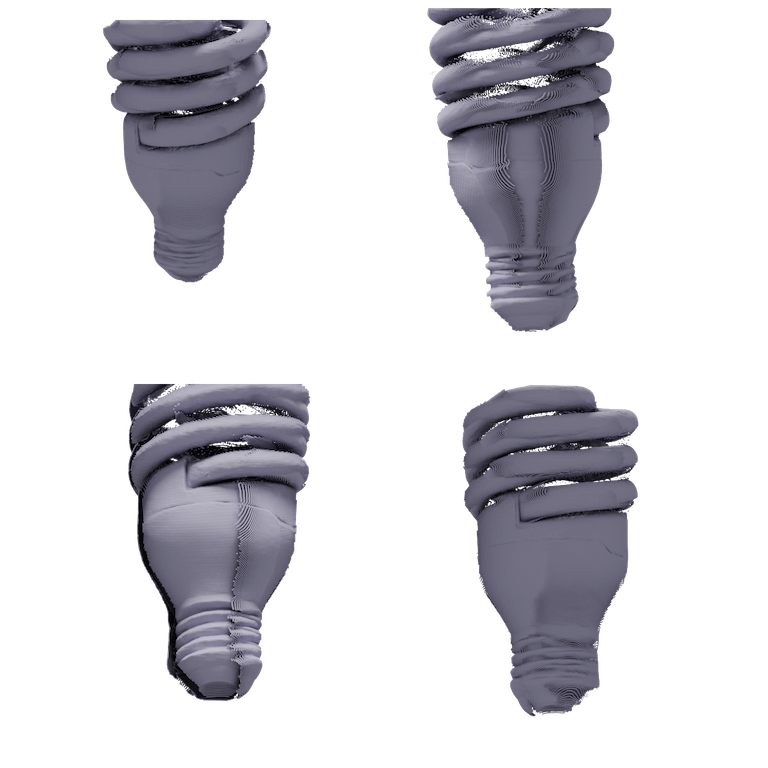} &
        \includegraphics[width=0.23\linewidth]{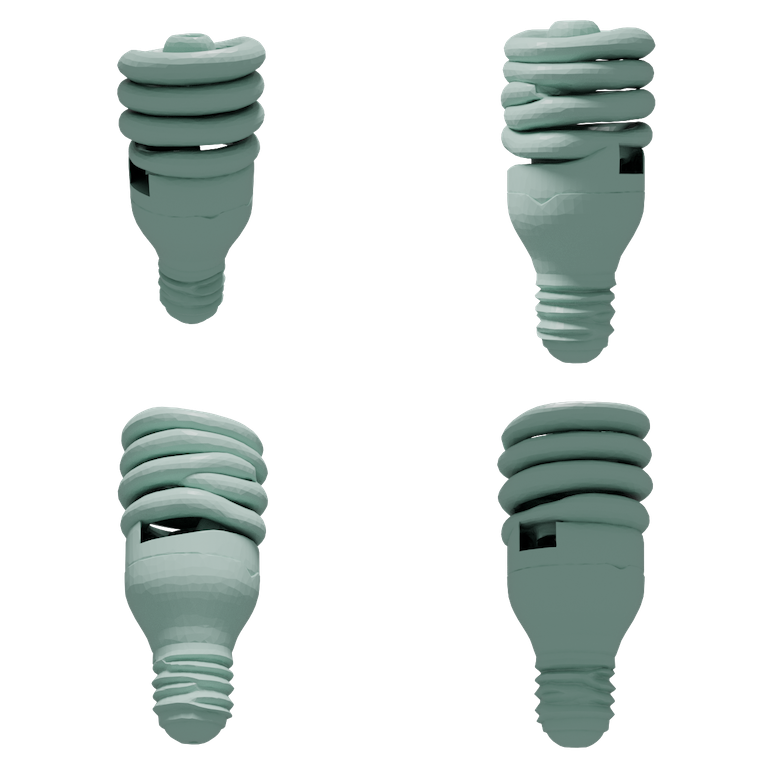} &
        \includegraphics[width=0.23\linewidth]{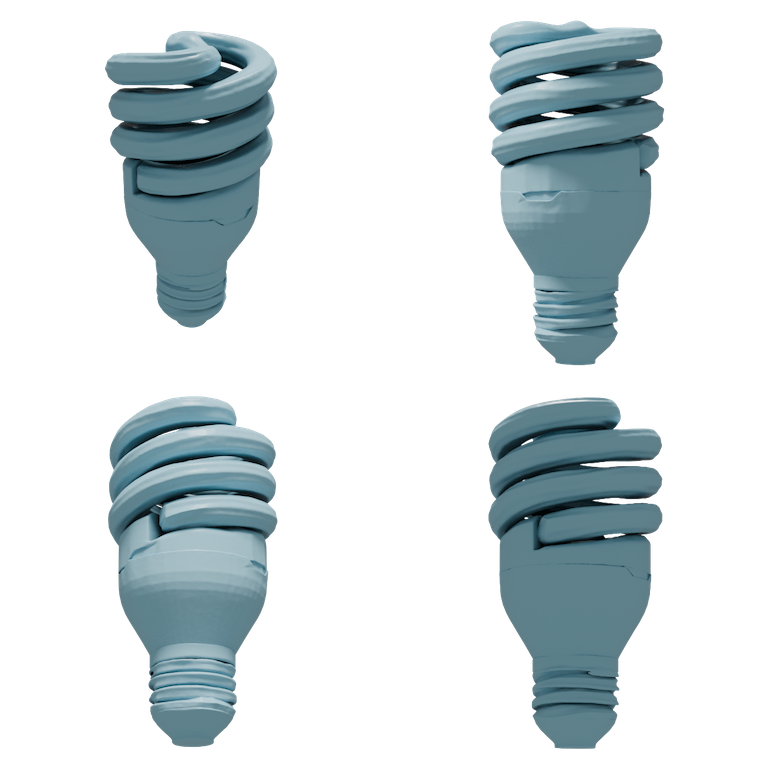} \\

        Input & VGGT & TRELLIS & Pixal3D  \\
    \end{tabular}

    \caption{Qualitative comparison of multi-view 3D generation on Toys4K.}
    \label{fig:multiview}
\end{figure*}

\begin{table*}[t]
\setlength{\tabcolsep}{4pt}
\centering
\caption{Quantitative evaluation of single-view generation on Toys4K. Metrics compare rendered and ground-truth normals.}
\vskip -0.1in
\label{tab:normal_metrics}
\begin{tabular}{l|cccc|cccccc}
\toprule
Method & IoU$\uparrow$ & PSNR$\uparrow$ &\revision{SSIM}$\uparrow$ & \revision{LPIPS}$\downarrow$ & Mean$\downarrow$ & Median$\downarrow$ & Mean\_B$\downarrow$ & ${11.25}^\circ$$\uparrow$ & ${22.5}^\circ$$\uparrow$ & ${30}^\circ$$\uparrow$ \\
\midrule
TRELLIS~\cite{xiang2025structured} & 79.48 & 20.98 &0.883 & 0.204 & 25.00 & 17.97 & 36.04 & 46.82 & 63.99 & 70.80\\
TripoSG~\cite{li2025triposg} & 73.54 & 19.73  &0.873 & 0.250& 28.55 & 21.20 & 41.71 & 39.85 & 57.18 & 64.81\\
Hunyuan3D-2.1~\cite{hunyuan3d2025hunyuan3d} & 83.33 & 21.96  &0.889 & 0.179& 21.19 & 14.05 & 32.46 & 51.37 & 69.08 & 75.83\\
Direct3D-S2~\cite{wu2025direct3d} & 74.23 & 19.49 &0.851 & 0.268 & 29.99 & 23.46 & 41.04 & 37.56 & 55.46 & 63.20\\
\midrule
\textbf{Pixal3D} & \textbf{93.57} & \textbf{24.21} &\textbf{0.897} & \textbf{0.108} & \textbf{16.63} & \textbf{11.77} & \textbf{21.80} & \textbf{53.13} & \textbf{77.96} & \textbf{85.35}\\
\bottomrule
\end{tabular}
\vskip -0.1in
\end{table*}%

\subsection{Scene Generation Pipeline}
\label{sec:scene}
Pixal3D enables pixel-aligned 3D object generation from images. Consequently, when the input is a scene image containing multiple objects, Pixal3D can generate each object individually and then compose them via image-space alignment, yielding a full 3D scene synthesis. A recent representative work for this task is SAM3D~\cite{sam3dteam2025sam3d3dfyimages}. SAM3D first leverages SAM3~\cite{carion2025sam} to interactively segment objects in image, and then trains an object generator on TRELLIS backbone. Given an RGB crop of a partially visible, occluded object, it generates the object in a canonical pose and predicts its pose (rotation, translation, and scale) in the camera frame to align objects into a consistent, object-separated 3D scene.

In contrast, Pixal3D generates each object directly in the camera space in a pixel-aligned manner, which makes multi-object alignment substantially simpler. We propose a modular scene generation pipeline comprising three steps: (1) \textit{Segmentation and Completion:} we employ SAM3~\cite{carion2025sam} for interactive segmentation to obtain object masks, followed by Qwen-image-edit~\cite{wu2025qwen} to perform 2D completion of occluded regions. (2) \textit{Pixel-Aligned Generation: }These completed images are fed into Pixal3D for 3D generation. Since the orientation of our generated objects is already aligned with the input image, we only need to resolve relative scale and depth across objects. (3) \textit{Global Alignment: }we use MoGe \cite{wang2025moge} to predict a global point map from image. The pixel-aligned nature of both Pixal3D's outputs and MoGe's predictions allow us to directly formulate point-wise constraints and solve a least-squares problem to estimate object scale and depth. 

Compared to SAM3D, this pipeline offers superior fidelity and geometric detail in single-object generation. Furthermore, our pipeline avoids the challenging and often non-robust step of estimating a 7-DoF object pose from the image. Instead, we resolve alignment through pixel-aligned generation and global depth estimation, which substantially improves both the accuracy and stability of multi-object alignment. Our pipeline yields higher-fidelity 3D scene generation results, providing a promising alternative perspective for holistic 3D scene generation from a single image.

\subsection{Implementation Details}
\label{sec:imple}
To train Pixal3D, we use the TRELLIS-500K~\cite{xiang2025structured} subset of Objaverse dataset~\cite{deitke2023objaversexl}. To construct pixel-aligned image-mesh pairs, we apply random object-centric rotations and render meshes from frontal perspectives with varying FoVs and camera distances. We watertight each mesh and compute its SDF. For model architecture, we use the same VAE and DiT model as Direct3D-S2~\cite{wu2025direct3d}, except that we replace the cross-attention conditioning with our back-projection-based conditioner. The pretrained VAE works robustly for pixel-aligned SDF, thus we only finetune the decoder for better quality. For sparse DiT, following Direct3D-S2, we adopt a coarse-to-fine training schedule, training at resolutions of 256, 384, 512, and 1024 for 200k, 100k, 80k, and 40k iterations, respectively. The dense DiT is trained for 300k iterations with a learning rate of 1e-4, followed by an additional 200k iterations at 2e-5. For image conditioning, we use DINOv2-Large as encoder, and employ NAF~\cite{chambon2025naf} to upsample to the DINOv2’s input resolution of 518x518. For multi-view training, the pre-trained single-view model is fine-tuned, with random sampling of 2 to 6 views as condition. Code will be released publicly.

\section{Experiments}
\begin{table}[t]
\setlength{\tabcolsep}{4pt}
\centering
\caption{Quantitative evaluation of single-view generation on in-the-wild test set. User study collects scores for both fidelity and quality.}
\vskip -0.1in
\label{tab:user_study}
\begin{tabular}{l|cc|cc}
\toprule
Method & Uni3D$\uparrow$ & ULIP2$\uparrow$ & Fidelity$\uparrow$ & Quality$\uparrow$\\
\midrule
TRELLIS & 41.09 & 44.76 & 1.86 & 1.99 \\
TripoSG & 40.99 & 44.64 & 2.25 & 2.14\\
Hunyuan3D-2.1 & 41.15 & 44.65 & 2.77 & 2.50\\
Direct3D-S2 & 41.62 & 44.79 & 3.21 & 3.64\\
\midrule
\textbf{Pixal3D} & \textbf{42.11} & \textbf{45.04} & \textbf{4.91}& \textbf{4.74} \\
\bottomrule
\end{tabular}
\vskip -0.15in
\end{table}%

\begin{table*}[t]
\centering
\caption{Quantitative evaluation of multi-view generation on Toys4K.}
\vskip -0.1in
\label{tab:multiview}
\scalebox{0.95}{
\begin{tabular}{l|ccc|ccc|ccc}
\toprule
% threshold=[0.0001, 0.0002]
\textbf{Method} & \multicolumn{3}{c|}{\textbf{View = 2}} & \multicolumn{3}{c|}{\textbf{View = 4}} & \multicolumn{3}{c}{\textbf{View = 6}} \\
 & CD ($10^{-4}$)$\downarrow$ & EMD ($10^{-2}$)$\downarrow$ & F-Score$\uparrow$ & CD ($10^{-4}$)$\downarrow$ & EMD ($10^{-2}$)$\downarrow$ & F-Score$\uparrow$ & CD ($10^{-4}$)$\downarrow$ & EMD ($10^{-2}$)$\downarrow$ & F-Score$\uparrow$ \\
\midrule
VGGT & 613.55 & 19.60 & 9.57 & 881.53 & 21.71 & 10.25 & 2791.10 & 25.33 & 9.67 \\
TRELLIS & 21.39 & 2.40 & 43.68 & 18.35 & 2.19 & 45.90 & 18.13 & 2.16 & 46.02 \\
\textbf{Pixal3D} & \textbf{5.27} & \textbf{1.13} & \textbf{64.94} & \textbf{4.73} & \textbf{1.05} & \textbf{67.85} & \textbf{4.16} & \textbf{1.00} & \textbf{69.04} \\
\bottomrule
\end{tabular}}
\vskip -0.1in
\end{table*}%

\subsection{Single-view 3D Generation}
To validate the effectiveness of our Pixal3D framework, we conduct comprehensive quantitative and qualitative evaluations against representative state-of-the-art 3D generation methods, including TRELLIS~\cite{xiang2025structured}, TripoSG~\cite{li2025triposg}, Hunyuan3D-2.1~\cite{hunyuan3d2025hunyuan3d}, and Direct3D-S2~\cite{wu2025direct3d}.

\paragraph{Quantitative Comparison} To precisely assess fidelity differences across methods, we render the surface normals of each generated 3D mesh in the input image coordinate frame, then compare these with ground-truth normal maps. This evaluation is performed on all meshes in the Toys4K dataset~\cite{stojanov2021using}. For the baselines, we use the ground-truth camera pose for normal rendering. In contrast, owing to its pixel-aligned nature, our method directly leverages its inference-time projection for normal rendering.

As for metrics, we use IoU to measure the overlap between the rendered and the ground-truth normal maps, and PSNR to quantify their pixel-wise discrepancy. In addition, we report commonly used error metrics in normal estimation: mean and median angular error, mean angular error around image boundaries (Mean\_B), and accuracy under different angular thresholds. All these metrics are computed only on the overlapping regions where both prediction and ground truth are available. The results are summarized in Table~\ref{tab:normal_metrics}. Our method achieves substantial improvements across all metrics, demonstrating significantly better fidelity.

Since the meshes in Toys4K are mostly simple, we further collect 150 images from the Internet and AI-generated sources as an additional test set, featuring complex geometric details and diverse semantics. On this test set, given the absence of ground-truth camera poses or normal maps, we evaluate image-3D consistency using ULIP2~\cite{xue2024ulip} and Uni3D~\cite{zhou2023uni3d}. We also conduct a user study with 30 participants on this test set. Participants are asked to score the generated meshes from two aspects: fidelity (image-3D consistency) and quality (overall 3D shape quality) from 1 (worst) to 5 (best). These results are summarized in Table~\ref{tab:user_study}. Our method is favored by the majority of participants, especially in terms of fidelity, highlighting its superior ability to faithfully preserve image details while maintaining high overall quality.

\paragraph{Qualitative comparisons} Figure~\ref{fig:singleiview_main} and~\ref{fig_only:singleiview} presents visual comparison examples. Compared to all other methods, our approach more faithfully and accurately recovers the visual content of input image and produces higher-quality 3D meshes. The fidelity gap is particularly evident in fine-grained details, such as keyboard layouts, facial details including eyes, the number and arrangement of flower petals, etc. These examples illustrate the misalignment caused by 2D-3D correspondence ambiguity in prior methods. In contrast, thanks to its pixel-aligned formulation, our method preserves nearly all image details, achieving an almost reconstruction-level fidelity. More examples are provided in the supplementary material and the video.

\subsection{Multi-view 3D Generation}
For multiview evaluation, we select representative baselines from both multiview reconstruction and generation, specifically VGGT \cite{wang2025vggt} and TRELLIS (multiview version) \cite{xiang2025structured}. Evaluations are conducted on the Toys4k dataset using Chamfer Distance (CD), Earth Mover’s Distance (EMD), and F-Score. We evaluate performance with varying numbers of input views (2, 4, and 6). Table~\ref{tab:multiview} summarizes the results.
Our method significantly outperforms the baseline approaches across all metrics. 

Qualitative results are presented in Figure~\ref{fig:multiview}. Specifically, VGGT often fails to produce strictly aligned point cloud reconstructions and frequently exhibits significant floaters and outliers. While the multi-view variant of TRELLIS produces smooth mesh outputs, its multi-view fidelity remains limited. It struggles to ensure consistency across all views and occasionally introduces hallucinations. In contrast, our pixel-aligned formulation seamlessly accommodates multi-view inputs, resulting in superior cross-view consistency. Moreover, as the number of views increases, generative ambiguity decreases while reconstruction cues become stronger, a trend consistently observed in our results. This behavior is also a fundamental principle and objective of 3D generative reconstruction.

\begin{figure}[tb]

    \centering
    \small
    \setlength{\tabcolsep}{0pt}
    \begin{tabular}{ccc}

        \centering
        \includegraphics[width=0.33\linewidth]{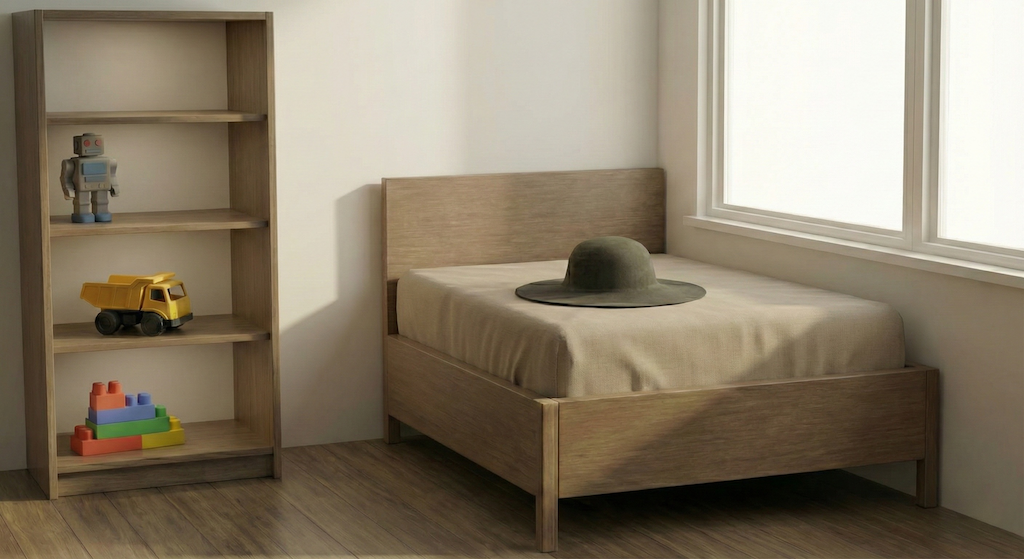} &
        \includegraphics[width=0.33\linewidth]{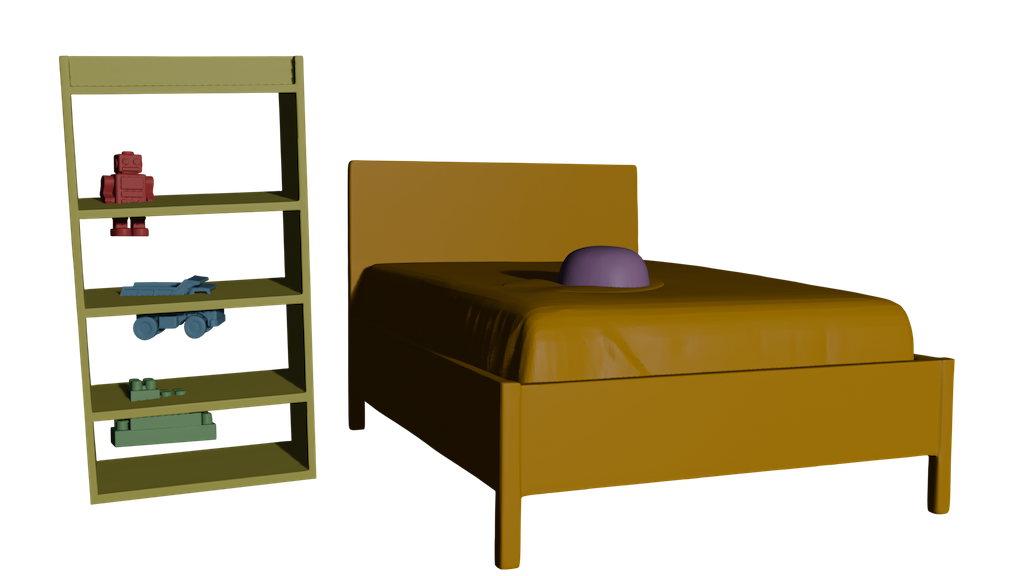} &
        \includegraphics[width=0.33\linewidth]{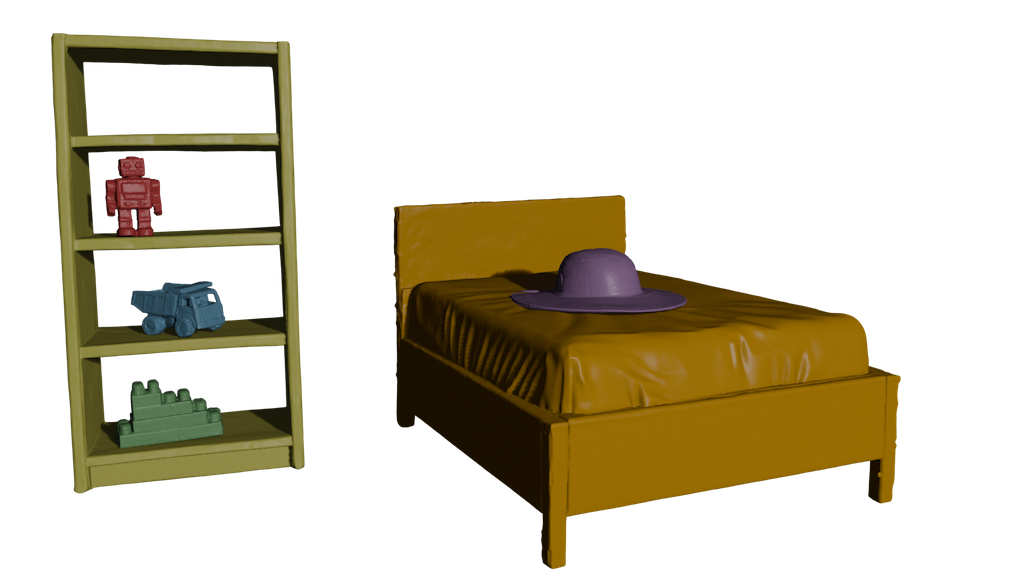} \\

       \includegraphics[width=0.33\linewidth]{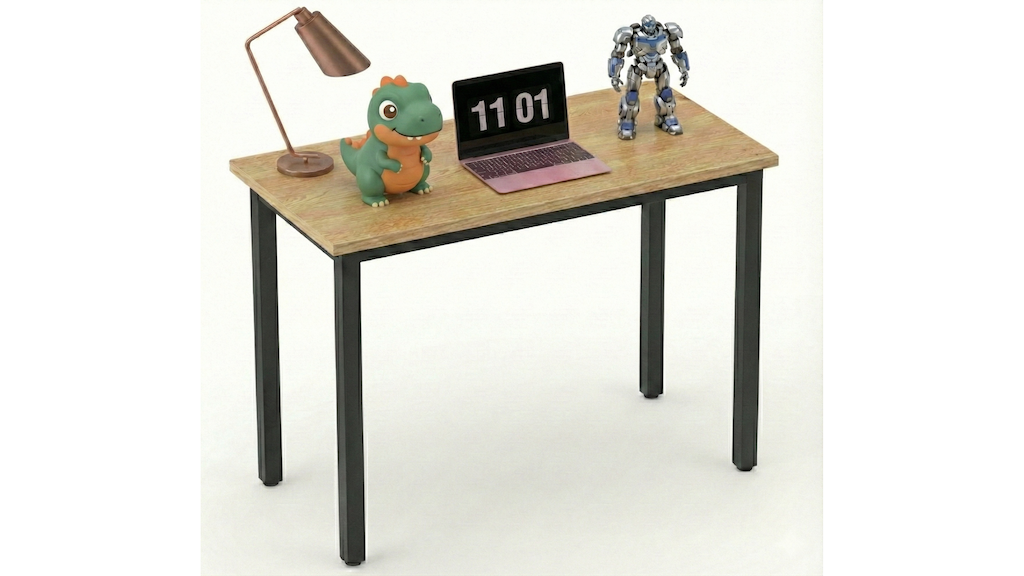} &
        \includegraphics[width=0.33\linewidth]{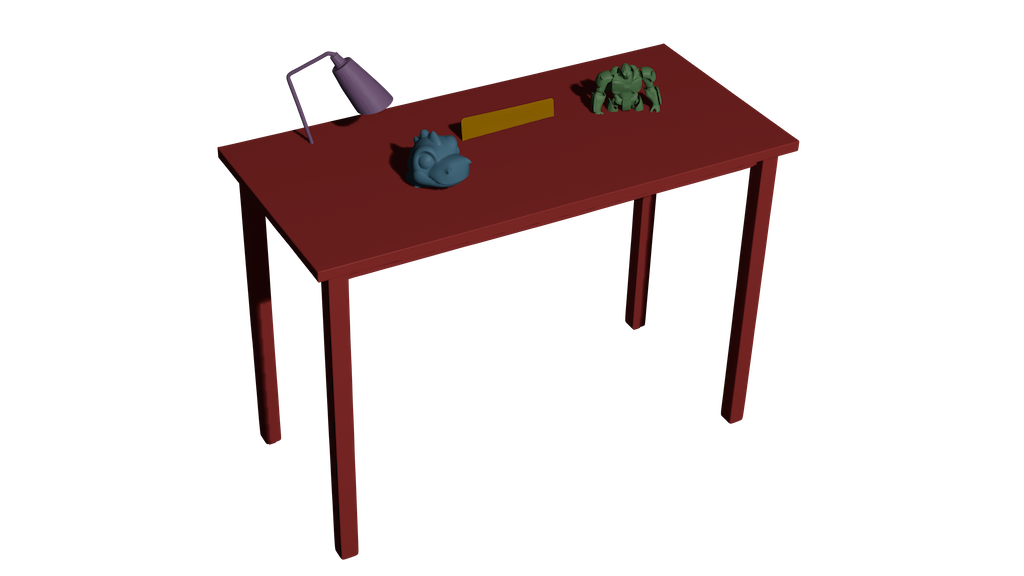} &
        \includegraphics[width=0.33\linewidth]{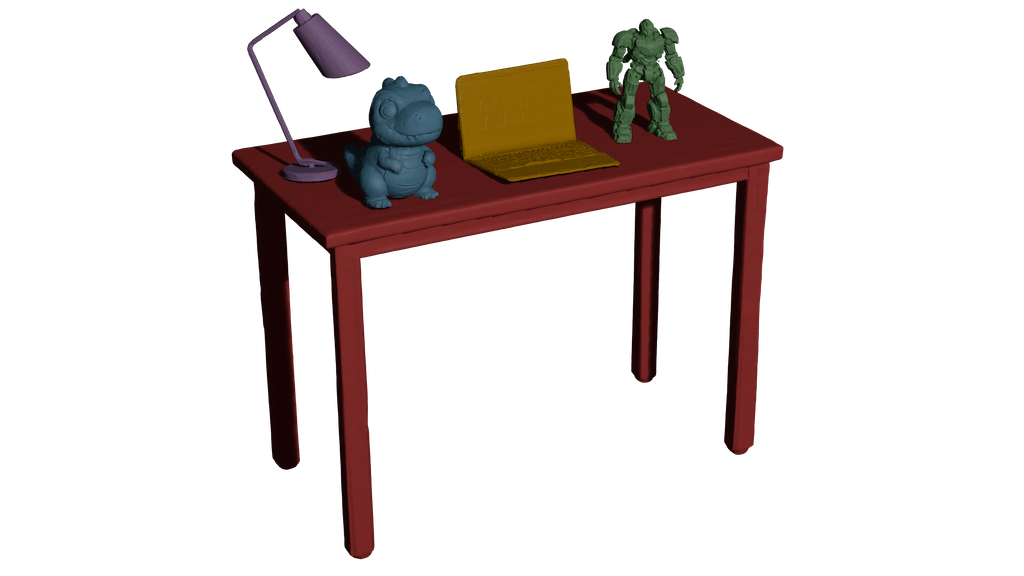} \\
       
       Input & SAM3D & Pixal3D  \\
    \end{tabular}
     \vskip -0.1in
  \caption{Qualitative comparison on 3D scene generation. 
  }
   
    \label{fig:scene}
    \vskip -0.1in
\end{figure}

\subsection{3D Scene Generation}
We extend Pixal3D to scene generation, as discussed in Sec.~\ref{sec:scene}. Figure~\ref{fig:scene} presents a qualitative comparison between our results and recent SAM3D~\cite{sam3dteam2025sam3d3dfyimages}. SAM3D jointly estimates canonical geometry and object poses, but its per-object pose estimation often yields inconsistent, non-robust inter-object relations (e.g., wrong relative rotations, misaligned placements, and incorrect contact/support), as shown in the figure. In contrast, our method enforces pixel-aligned constraints for each object relative to the input image and regularizes their spatial consistency through geometric cues like depth maps, leading to more coherent and practically usable scene generation results. 
\begin{figure}[tb] 
    \centering
    \includegraphics[width=0.98\linewidth]{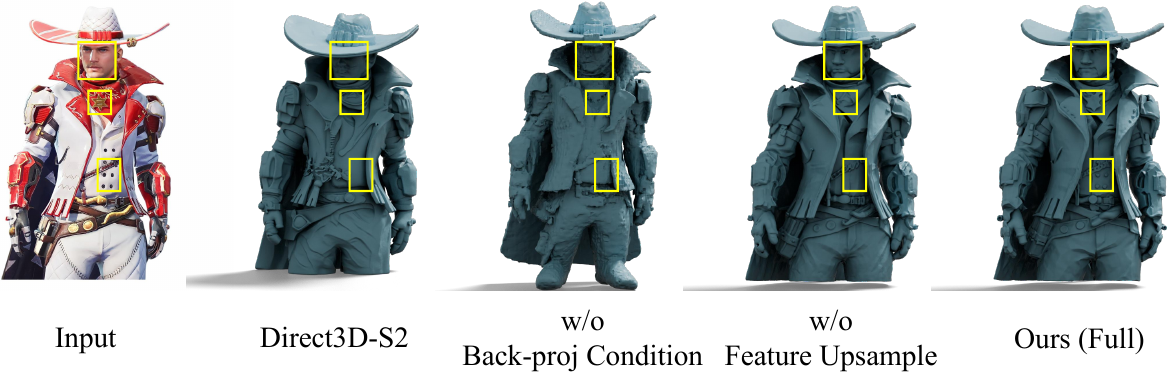} 
    \vskip -0.1in
    \caption{Ablation study on key components. }
    \label{fig:ablation}
    \vspace{-10pt}
\end{figure}

\subsection{Ablation Studies}
We conduct ablation studies to validate the effectiveness of our key modules. The results are shown in Figure~\ref{fig:ablation}. Without feature upsampling, 3D generation must rely on relatively coarse feature maps (e.g., 37$\times$37 patch tokens from DINOv2) to represent image details. This inevitably leads to missing fine details and misalignment in the generated 3D results, as illustrated in the figure. Furthermore, when the back-projection conditioning scheme is removed and replaced by a conventional cross-attention mechanism for a pixel-aligned 3D generator, training becomes slow to converge and unstable, and the final results exhibit substantially lower fidelity. These observations highlight the necessity of our design choices.

\subsection{Limitations and Future Works}
While our method demonstrates strong 3D generative reconstruction performance, several limitations remain. First, the framework exhibits sensitivity to pixel-level noise (e.g. imperfect segmentation boundaries), which can be back-projected and amplified into small geometric artifacts. Second, our current multi-view formulation assumes known and reasonably accurate camera poses. Third, our scene-generation pipeline relies on 2D inpainting to complete occluded regions, which may occasionally introduce errors in complex occlusions. Moving forward, a natural next step is to extend the current geometry backbone with texture and material synthesis, where our pixel-aligned paradigm is particularly suited to improve appearance fidelity. In addition, pixel-aligned generation opens opportunities for downstream 3D editing and interaction via 2D pixel manipulation. Finally, extending pixel-aligned generation to video-based 3D scene generation would be an interesting direction, bridging high-fidelity asset creation with controllable world building.

\section{Conclusion}
In this paper, we present Pixal3D, a pixel-aligned 3D generation paradigm for high-fidelity 3D asset creation from images. Unlike existing 3D-native generation methods that synthesize shapes in canonical space, Pixal3D directly creates 3D models aligned with images. A back-projection based image condition scheme replaces ambiguous cross-attention with explicit, geometric 2D-3D correspondence, enabling high-precision, pixel-aligned synthesis. We further demonstrated the versatility of this paradigm by extending it to multi-view inputs and 3D scene generation through a modular pipeline. Our extensive evaluations confirm that pixel-aligned generation is not only feasible but significantly enhances 3D fidelity. Pixal3D provides a scalable foundation for 3D generative reconstruction, offering a promising path towards creating 3D content that is both creatively flexible and pixel-faithful.

\begin{acks}
This work was supported by Fundamental and Interdisciplinary Disciplines Breakthrough Plan of the Ministry of Education of China (No. JYB2025XDXM101), the National Natural Science Foundation of China (project No. 62495060), the Research Grant of Tsinghua-Tencent Joint Laboratory for Internet Innovation Technology.
\end{acks}

% Bibliography
\bibliographystyle{ACM-Reference-Format}
\bibliography{bib}

\end{document}